\documentclass[journal]{IEEEtran}

\usepackage{makecell}
\newcommand{\PreserveBackslash}[1]{\let\temp=\\#1\let\\=\temp}
\newcolumntype{C}[1]{>{\PreserveBackslash\centering}p{#1}}
\newcolumntype{R}[1]{>{\PreserveBackslash\raggedleft}p{#1}}
\newcolumntype{L}[1]{>{\PreserveBackslash\raggedright}p{#1}}

\usepackage{graphicx}
\usepackage{amsmath,amssymb} 
\usepackage{multirow}
\usepackage{array}

\usepackage{times}
\usepackage{epsfig}
\usepackage{mathrsfs}
\usepackage{multirow}
\usepackage[labelsep=period]{caption}
\usepackage{subfig}
\usepackage{url}
\usepackage{cite}
\usepackage{color}

\ifCLASSINFOpdf
 
\else
  
\fi

\begin{document}
\title{IAUnet: Global Context-Aware Feature Learning for Person Re-Identification}

\author{Ruibing~Hou,~\IEEEmembership{Student Member,~IEEE}, Bingpeng~Ma, Hong~Chang,~\IEEEmembership{Member,~IEEE}, Xinqian~Gu,~\IEEEmembership{Student Member,~IEEE}, Shiguang~Shan,~\IEEEmembership{Senior~Member,~IEEE}, Xilin~Chen,~\IEEEmembership{Fellow,~IEEE}
\IEEEcompsocitemizethanks{\IEEEcompsocthanksitem R. Hou, H. Chang, X. Gu and X. Chen are with Key Lab of Intelligent Information Processing of Chinese Academy of Sciences (CAS), Institute of Computing Technology, CAS, Beijing, 100190, China and University of Chinese Academy of Sciences, Beijing 100049, China.

B. Ma is with University of Chinese Academy of Sciences, Beijing 100049, China.

S. Shan is with Key Lab of Intelligent Information Processing of Chinese Academy of Sciences (CAS), Institute of Computing Technology, CAS, Beijing, 100190, China and University of Chinese Academy of Sciences, Beijing 100049, China and CAS Center for Excellence in Brain Science and Intelligence Technology, Shanghai, 200031, China. \protect\\

E-mail: \{ruibing.hou, xinqian.gu\}@vipl.ict.ac.cn, bpma@ucas.ac.cn, \{changhong, sgshan, xlchen\}@ict.ac.cn
}}

\markboth{Journal of \LaTeX\ Class Files,~Vol.~14, No.~8, August~2015}%
{Shell \MakeLowercase{\textit{et al.}}: Bare Demo of IEEEtran.cls for IEEE Journals}

\maketitle

\IEEEdisplaynontitleabstractindextext
\IEEEpeerreviewmaketitle
\begin{abstract}
Person re-identification (reID) by CNNs based networks has achieved favorable performance in recent years. However, most of existing CNNs based methods do not take full advantage of spatial-temporal context modeling. In fact, the global spatial-temporal context can greatly clarify local distractions to enhance the target feature representation. To comprehensively leverage the spatial-temporal context information, in this work, we present a novel block, \textit{Interaction-Aggregation-Update} (IAU), for high-performance person reID. Firstly, \textit{Spatial-Temporal} IAU (STIAU) module is introduced. STIAU jointly incorporates two types of contextual interactions into a CNN framework for target feature learning. Here the spatial interactions learn to compute the contextual dependencies between different body parts of a single frame. While the temporal interactions are used to capture the contextual dependencies between the same body parts across all frames. Furthermore, a \textit{Channel} IAU (CIAU) module is designed to model the semantic contextual interactions between channel features to enhance the feature representation, especially for small-scale visual cues and body parts. Therefore, the IAU block enables the feature to incorporate the globally spatial, temporal, and channel context.
It is lightweight, end-to-end trainable, and can be easily plugged into existing CNNs to form IAUnet. The experiments show that IAUnet performs favorably against state-of-the-art on both image and video reID tasks and achieves compelling results on a general object categorization task. The source code is available at {\color{red}\url{https://github.com/blue-blue272/ImgReID-IAnet}}.
\end{abstract}
\begin{IEEEkeywords}
Person Re-Identification, Spatial-Temporal Context Modeling, Feature Enhancing, Interaction-Aggregation
\end{IEEEkeywords}

\IEEEpeerreviewmaketitle

\section{Introduction}
\label{sec:introduction}

Person re-identification (reID) aims at retrieving particular persons from non-overlapping camera views. It has gained increasing attention due to its importance in many applications, such as video surveillance analysis and tracking. Despite much progress has been achieved in recent years~\cite{multi-channel,context-aware,harmoniou,Hydraplus-net,background,smoothed}, it remains difficult due to tremendous challenges such as occlusion, background clutters, poses variation, and camera viewpoints variations. Previous approaches mostly focus on image setting, where the persons from different cameras are matched by comparing their still images. With the emergence of video benchmarks~\cite{mars,dukereid}, the researchers have also started to utilize video data for reID.

Recently, reID by deep neural networks has attracted increasing attention. These approaches utilize Convolutional Neural Networks (CNNs), which typically stack convolutional and pooling operations, to develop  discriminative and robust features.
With the powerful deep networks and large-scale labeled datasets, CNNs based methods achieve favorable performance and efficiency.

\begin{figure}[t]
\centering
   \includegraphics[width=1.0\linewidth]{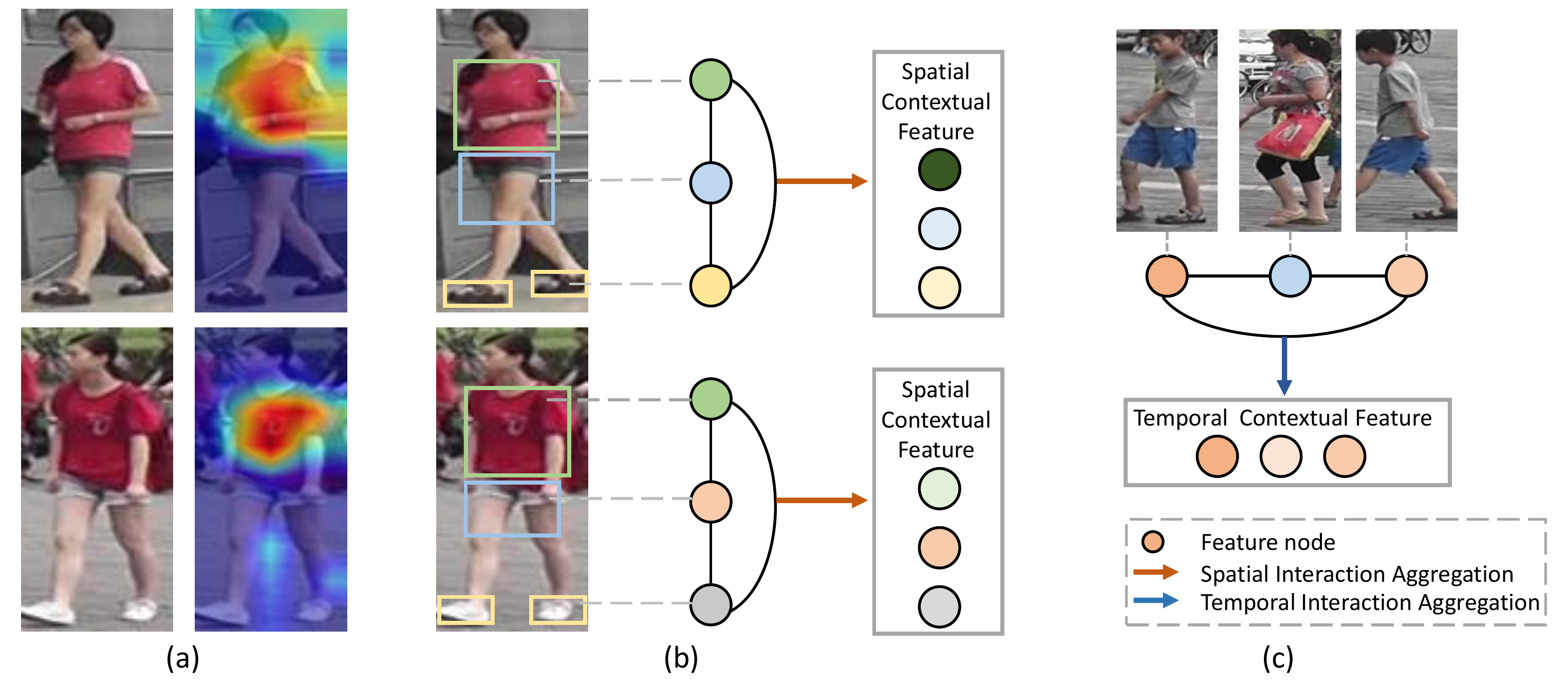}
   \caption{Illustration of our motivation. (a) The pair of input images and activation maps~\cite{zhou2016learning}. The upper clothes attract the most attention. But they are indistinguishable for the two persons. (b) In spatial context modeling, different body parts interact and aggregate to form structure features with spatial contextual knowledge. Here the upper body feature can be adaptively updated to distinguish the two persons. (c) In temporal context modeling, the frames interact and aggregate to generate features with temporal contextual information. With the temporal context, the corruption from the mis-detected frames can be alleviated.}
\label{fig1}
\vspace*{-1em}
\end{figure}

Despite the significant progress in image person reID, most existing CNNs based methods do not take full advantage of spatial context modeling. As point out by~\cite{zheng2019re}, the final convolutional features of pedestrians usually focus only on the most representative local regions, which may be indistinguishable for two persons with similar looking local parts. For example, as shown in Fig.~\ref{fig1}(a), the upper clothes of the image pair attract the most attention. But it is difficult to distinguish the two pedestrians. Varior \textsl{et al.}~\cite{short-term} demonstrate that the long-range \textbf{global spatial context} can greatly help to clarify local confusions. As illustrated in Fig.~\ref{fig1} (b), with the help of the spatial context, the features of upper body parts can be adaptively changed to distinguish the two pedestrians. Therefore, it is desirable to automatically capture the global spatial context for image reID.

For video person reID, current CNNs based methods do not make the best of spatial-temporal context modeling. The 2D convolution operations completely ignore the temporal information of the video. Although the 3D convolution~\cite{3D-convolution} operations can capture spatial-temporal context, they are limited to local temporal context modeling~\cite{non-local}. With only the spatial information, the feature generated for a video is often corrupted by the mis-detected frames~\cite{diversity}. McLaughlin \textsl{et al.}~\cite{RCN} point out that the long-range \textbf{global temporal context} can help to reduce the interference. As shown in Fig.~\ref{fig1} (c), with the help of temporal context, the features of the mis-detected frames can be adaptively updated to describe the target person. Therefore, it is necessary to capture the long-range spatial-temporal context for video reID.

A recent work~\cite{IANet} proposes an Interaction-and-Aggregation (IA) network for \textbf{image reID}. It introduces two modules that aid CNNs in modeling contextual dependencies. One is spatial IA, which models the dependencies between the features of fixed space positions and then aggregates the correlated features belonging to the same body parts. The other is channel IA, where the channel features with similar semantics are aggregated. The incorporation of these modules into a network gives it the ability to adapt its feature representation to contain contextual information.

Following~\cite{IANet}, we further propose a unified framework for \textbf{image and video reID}. Compared to~\cite{IANet}, the major changes of methods are two-fold. For one thing, different from~\cite{IANet} which models the spatial dependencies between fixed geometric positions, we go one step further and perform higher-level context modeling between disjoint and distant body parts regardless of their shapes. This is conductive to capture longer-term spatial contextual dependencies. For another, video reID is considered. We design an additional network module to capture the longer-range temporal contexts which can achieve more robust video feature representations.

In this work, we propose a new network module, \textit{Spatial-Temporal Interaction-Aggregation-Update} (STIAU), to jointly consider both globally spatial and temporal contexts of the target person. To be specific, given a video feature map, a sequence of intermediate convolutional feature maps of all frames, STIAU generates a spatial-temporal relation map. The relation map captures two types of interactions between body parts: spatial interactions which model the dependencies between disjoint and distant body parts of a single frame, and temporal interactions which model the dependencies between the body parts with the same semantics across all frames. In this way, the long-range spatial-temporal context of the video is captured. Based on the relation map, the features of different parts across all video frames are aggregated to generate a spatial-temporal contextual representation. Finally, the spatial-temporal context is incorporated into each frame to form a structured spatial-temporal feature. 

Similar to STIAU in principle, the \textit{Channel Interaction-Aggregation-Update} (CIAU) is proposed to further enhance the feature representation via modeling the semantic contextual interactions between the channels of the video feature maps. Specially, for small-scale body parts that easily fade away in the high-level features from CNNs, CIAU can selectively aggregate the semantically similar features across all channels to update and manifest their feature representations.

Both modules are computationally lightweight and impose only a slight increase in model complexity. They can be integrated into an Interaction-Aggregation-Update (IAU) block and readily inserted into CNNs at any depth. In this work, we add IAU block to ResNet-50~\cite{residual} to generate Interaction-Aggregation-Update Network (IAUnet) for person reID. To demonstrate the universality of IAU block, we also present results beyond ResNet-50, indicating that the proposed modules are not restricted to specific network architecture.

The contributions of this paper are summarized as follows: (1) We propose an unified network, IAUnet, for both image and video person reID. (2) We formulate an STIAU block for learning context-aware features. It designs the \textbf{interaction and aggregation} operations which can efficiently capture the \textbf{long-range and global context}. (3) We propose a CIAU block to model the contextual interactions  between feature channels. It can further enhance the feature representation by aggregating the semantically similar features. To our knowledge, we are the first to jointly exploit the spatial, temporal, and channel contexts in reID. Experiments on five reID benchmarks show the superiority of the proposed approach. Moreover, IAUnet is effective on general object categorization tasks as demonstrated on CIFAR-100~\cite{cifar100}, showing its potential beyond person reID.

\section{Related Work}
\label{section:relatedwork}
\subsection{Image Person ReID}
Image person reID has a very rich literature and can be divided into two classes: traditional and deep learning based approaches. Traditional solutions generally have two stages: extracting hand-crafted features and designing robust metrics. Various hand-crafted features have been developed. For metric learning, lots of metric learning techniques have been designed to decide whether two images are matched or not. On the other hand, the success of deep learning in image classification has been inspiring a lot of studies in image person reID \cite{background,smoothed,Learning,stepwise,hard-aware,gu2019temporal} \cite{ShiEmbeding,Yang2016Large,Rui2013}. A line of the work uses the siamese network which takes image pairs or triplets as the inputs. Li \textsl{et al.}~\cite{Cuhk} input a pair of pedestrian images to a CNN and the model is trained with a verification loss. Hermans \textsl{et al.}~\cite{Triplet} further employ a triplet loss. Another line adopts identity classification models. Further, Zhang \textsl{et al.}~\cite{zhang2019densely} train the model with a joint triplet and classification loss, which achieves the state-of-the-art performance.

\textbf{Spatial Context Modeling.} To handle various challenges in person reID, several algorithms have been proposed to impose spatial structure information on target person appearance modeling. The part-based methods that decompose the target person into several parts have been studied actively. For example, the human parsing method~\cite{semantic,mask-guided}, pose detection method~\cite{spindle-net,pose-invariant}, and body part specific attention modeling~\cite{part-aligned,PCB,IANet} have been designed to localize body parts for part-aligned feature extracting and matching. However, the part-based methods ignore the spatial context between different parts, thus the similar looking local parts may lead to wrong retrieval results. Recently, Varior \textsl{et al.}~\cite{short-term} employ Long Short-Term Memory (LSTM) to model the spatial correlations between different local parts. The work demonstrates that the spatial contextual information is beneficial to enhance the discriminative capability of local features. However, LSTM can not explicitly model the interactions between local body parts. It causes optimization difficulties that need to be carefully addressed~\cite{fd}. In contrast, the proposed IAU block is lightweight and simple, specialized to model global context in a computationally efficient manner and designed to enhance the discriminative power of features.
\begin{figure*}[t]
\centering
   \includegraphics[width=1.0\linewidth]{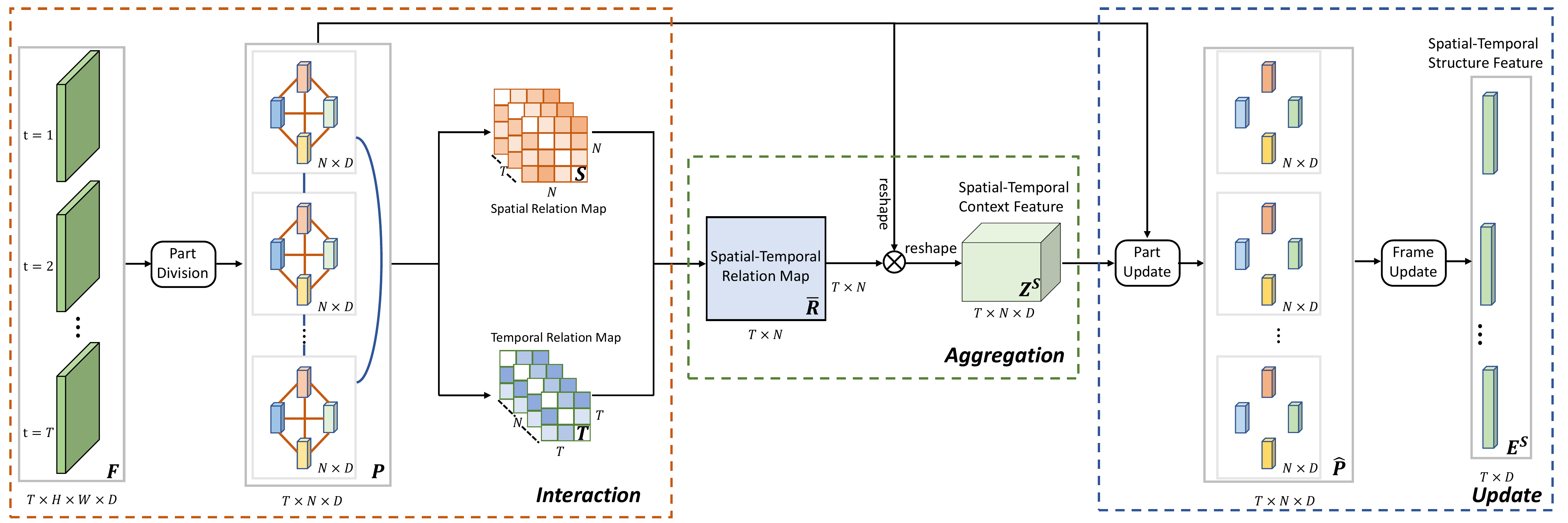}
   \caption{The architecture of STIAU module.}
\label{STIAU}
\vspace*{-1em}
\end{figure*}

\subsection{Video person reID.}

Video person reID is an extension of image approaches, where the sequential data is used instead of an individual image. Early video reID methods mainly focus on handcrafting video features. The powerful feature learning ability of CNN also inspires its application in video reID. To better distill discriminative information from video, the attention based approaches are gaining popularity. Liu \textsl{et al.}~\cite{QAN} predict a quality score for each frame to weaken the influence of noisy samples. Zhou \textsl{et al.}~\cite{See} propose a RNN based attention mechanism to select the most discriminative frames from the video. Further, the works~\cite{jointly,diversity} employ a spatial and temporal attention layer. However, the attention based methods usually discard the disturbed frames directly, resulting in the loss of spatial and temporal information of video data. In contrast, we explicitly utilize the spatial-temporal context to alleviate the influence of the disturbed frames, without losing any spatial-temporal information.

\textbf{Spatial-Temporal Context Modeling.} Recently, many methods~\cite{li2019multi,gu2019temporal,HouTCL,Gu3D} propose to exploit temporal context on target sequence appearance modeling. A branch of works~\cite{RCN,snippet} uses the optical flow that provides the motion features. But the optical flow only captures the local temporal context between adjacent frames. Another branch adopts the RNN~\cite{jointly,RCN,See} to explore the long-range temporal context. Nevertheless, they can only capture the temporal contextual relations in the end. Thus they can not build a hierarchical structure. Besides, all the above methods ignore the spatial context of videos. On the contrary, the proposed IAU block captures both spatial and temporal contextual information, and can be added to the earlier part of CNNs to build a richer hierarchy.

\subsection{Fine-grained Visual Categorization.}
Fine-grained visual categorization aims to discriminate similar subcategories that belong to the same superclass. Since the distinctions among similar subcategories are quite subtle and local, existing methods~\cite{Gaze,OPA,WSFG,WSL} usually adopt the features of local parts to represent the images. For example, He \textsl{et al.}~\cite{Gaze} utilize object and part detectors to extract part features, which is free of using object and part annotations. Zhang \textsl{et al.}~\cite{OPA} propose to use a weakly supervised method to generate the part proposals. Further, the works~\cite{WSFG,WSL} propose a weakly supervised part selection method with spatial constraints. In this work, we also use a weakly supervised part division unit to extract the body part features for input images. However, different from the above works, the proposed method further models the relations between different parts. This can greatly help to clarify the local confusion caused by seemingly alike parts of different pedestrians.

\subsection{Spatial-Temporal Context Modeling.}
Generalization of neural networks to automatically model spatial-temporal contextual relations has drawn great attention recently. Chen \textsl{et al.}~\cite{convlstm} propose a convolutional LSTM for spatial-temporal sequence forecasting. Ji~\textsl{et al.}~\cite{3D-convolution} develop a 3D convolution to capture the motion information for video action recognition. To model long-range dependencies, Wang \textsl{et al.}~\cite{non-local} propose a non-local network to model the similarity relations between any pairs of positions. Wang \textsl{et al.}~\cite{wang2018videos} further capture the location and similarity relations between the detected objects for video recognition. Gao~\textsl{et al.}~\cite{GCT-Gao} learn a fixed temporal relation between frames to update the exemplar image for visual tracking. However, the above methods usually model the contextual relations between fixed geometric positions. In the proposed approach, we go one step further and perform higher-level spatial-temporal contextual modeling between disjoint and distant \textbf{body parts} regardless of their shape. Comprehensive empirical results verify the effectiveness of the proposed method.

\section{The proposed Method}
In this section, we first introduce STIAU and CIAU modules. Then, the IAU block, which integrates the two modules, is illustrated. Finally, we present the overall IAUnet for image and video person reID.

\subsection{STIAU Module}
Spatial-temporal context of the target person sequence is crucial for video person reID. However, most existing methods either lack the ability of modeling long-range spatial contextual relationships or overlook the temporal contextual knowledge, resulting in highly sensitive to the distracting objects. To this end, we design an STIAU module to form a structured representation with the spatial-temporal context of the target person sequence.

As shown in Fig.~\ref{STIAU}, suppose a convolutional video feature map $F\in\mathbb{R}^{T\times H\times W \times D}$ is given, where $T$, $H$, $W$, and $D$ denote the frame number, the height, the width, and the channel number of the feature map respectively. We first use a part division unit to extract the part features for each frame. The part features are associated with different body regions, namely head, upper-body, lower-body, and shoes. Then, we feed the part features into three sequential operations, \textit{interaction}, \textit{aggregation}, and \textit{update}. \textit{Interaction} operation explicitly models the dependencies between the parts to produce a spatial-temporal relation map. Two types of relations are considered: spatial relations and temporal relations. The generated relation map is then used to aggregate correlated part features in the following \textit{aggregation} operation, producing a spatial-temporal context feature. Finally, the context feature is utilized in the \textit{update} operation to obtain the feature with spatial-temporal structure information.

\textbf{Interaction Operation.}
 As illustrated in Fig.~\ref{STIAU}, the part division unit takes the video feature map $F$ as inputs and produces the corresponding video part features $P\in\mathbb{R}^{T\times N\times D}$, where $N$ is the number of the parts of each frame. The details of the part division unit will be described later. To perform spatial-temporal context modeling of the target person sequence, the \textit{interaction} operation models the global contextual relations between the $N$ parts across the $T$ frames, featuring both spatial and temporal relationships.

 Specifically, we set $P=\{p_{ij}|i=1,\dots,T,j=1,\dots,N\}$ consisting of all body parts in a person sequence, where $p_{ij}\in\mathbb{R}^{D}$ is the $j^{th}$ body part feature of the $i^{th}$ frame. Two types of contextual relations are considered, spatial relations and temporal relations. The spatial relations represent the part interactions within the frames. To be specific, we model the spatial relations between any pairs of body parts of each frame, producing a spatial relation map $S=\{S_i\}_{i=1}^{T}$. Here $S_i\in\mathbb{R}^{N\times N}$ is the spatial relation map of the ${i}^{th}$ frame, which is defined as:
 \begin{equation}
 \left(S_{i}\right)_{jk}=W_r^T\left(\left[|p_{ij}-p_{ik}|, u\right]\right),
 \text{where}, u=\textit{GAP}(F), k \neq j.
 \label{eq1}
\end{equation}
Here $|.|$ denotes the absolute value, $[.,.]$ denotes concatenation, and $W_r$ is a weight vector that projects the concatenated vector to a relation scalar. $\textit{GAP}$ stands for Global Average Pooling operation. It performs global spatial-temporal average pooling to the video feature map $F$ to form a coarse global feature of the video, denoted as $u$ ($u\in\mathbb{R}^{D}$). With the global feature $u$, the local relations between body parts can be estimated in a global view.

The temporal relations represent the part interactions among frames. In particular, we model the temporal relations of the body parts with the same semantic across all frames. It generates a temporal relation map $T=\{T_i\}_{i=1}^{N}$. Here $T_i\in\mathbb{R}^{T\times T}$ is the temporal relation map of the $i^{th}$ body part, which is denoted as:
\begin{equation}
 \left(T_{i}\right)_{jk}=W_r^T\left([|p_{ji}-p_{ki}|, u]\right), \text{where}, k\neq j.
 \label{eq2}
\end{equation}
As shown in Eq.~\ref{eq2}, the coarse global feature $u$ is also used as an input to predict the temporal relations.

In the last, we integrate $S$ and $T$ to form the spatial-temporal relation map $R\in\mathbb{R}^{TN\times TN}$:
\begin{equation}
R_{ij}=\left \{
\begin{array}{lcl}
\left(S_{t_1}\right)_{n_1n_2} & & {t_1=t_2} \\
\left(T_{n_1}\right)_{t_1t_2} & & {t_1 \neq t_2\ \text{and} \ n_1=n_2}  \\
0 & & {t_1 \neq t_2 \ \text{and} \ n_1 \neq n_2}  \\
\end{array} \right.,
 \label{eq3}
\end{equation}
where $t_1=i/N+1$, $n_1=i(mod \ N)+1$, $t_2=j/N+1$, $n_2=j(mod \ N)+1$, and $R_{ij}$ denotes the relations between the $n_1^{th}$ part of the $t_1^{th}$ frame and the $n_2^{th}$ part of the $t_2^{th}$ frame. A modified softmax is then used to normalize the relation map:
\begin{equation}
\overline{R}_{ij} = \left \{
\begin{array}{lcl}
 \frac{\exp\left(R_{ij}\right)}{\sum_{k,R_{ik}\neq 0} \exp\left(R_{ik}\right)} & & {R_{ij} \neq 0}\\
 0 & & {R_{ij}=0} \\
 \end{array} \right..
\end{equation}
Notably, with the decomposition of spatial and temporal relations, each body part is related to $N+T-1$ parts among a total of $NT$ input parts, which significantly reduces the computation cost of the \textit{interaction} operation.

\textbf{Aggregation Operation.}
To make use of $\overline{R}$ in the \textit{interaction} operation, we follow it with the \textit{aggregation} operation which aims to aggregate the input video part features based on the relation map. As shown in Fig.~\ref{fig1}, we first reshape $P$ to $R^{TN\times D}$, and then perform the matrix multiplication between $\overline{R}$ and $P$ to obtain the spatial temporal context feature $Z^S\in\mathbb{R}^{TN\times D}$:
\begin{equation}
Z^S=\overline{R}P.
\end{equation}
$Z^S$ is then reshaped to $\mathbb{R}^{T\times N\times D}$ to maintain the size of the input video part feature.

\textbf{Update Operation.}
With the spatial-temporal context feature, we can compute updated part features using a \textit{part update unit}. It fuses the initial part feature $P$ and part context feature $Z^S$ to produce the adapted feature $\hat{P}\in\mathbb{R}^{T\times N\times D}$:
\begin{equation}
\hat{p}_{ij}=W_{pu}^T\left([p_{ij}, z^S_{ij}]\right),
\end{equation}
where $\hat{p}_{ij}\in\mathbb{R}^{D}$ is the $j^{th}$ part feature of the $i^{th}$ frame of $\hat{P}$, and analogously for $z^S_{ij}$, and $W_{pu} \in \mathbb{R}^{2D\times D}$ computes per-part update on the concatenated vector and maintains the input feature dimensionality.

Finally, a \textit{frame update unit} is applied to each frame to obtain the spatial-temporal structure feature. It firstly performs global average pooling to $\{\hat{P_i}\}_{i=1}^{T}$ to generate the global feature for each frame. Then it integrates the frame global feature with the coarse video global feature $u$ to form the spatial-temporal structure feature $E^S\in\mathbb{R}^{T\times D}$:
\begin{equation}
E^S_i = W_{fu}^T\left(\left[\frac{\sum_j \hat{p}_{ij}}{N}, u\right]\right),
\label{eq7}
\end{equation}
where $W_{fu} \in \mathbb{R}^{2D\times D}$ computes per-frame update on the concatenated vector and maintains the input feature dimensionality.

\textbf{Part Division Unit.}
To exploit the local part features for the STIAU module, we should firstly localize the regions of different body parts. Existing methods~\cite{semantic,part-aligned,zhang2019densely,PNGAN} usually utilize an external part detection network, making the reID framework too complicated and time consuming. In contrast, we adopt a simple and lightweight spatial attention subnet to localize the body parts. Specifically, taking the video feature $F$ as inputs, the subnet uses a convolutional layer to produce the attention maps $A\in\mathbb{R}^{T \times H \times W\times N}$ associated with different body parts:
\begin{equation}
A = \sigma \left(W_a  * F + b_a\right),
\end{equation}
where $\sigma$ denotes the sigmoid function, $*$ is the convolutional operation, and $W_a\in\mathbb{R}^{1\times 1\times D \times N}$ and $b_a\in\mathbb{R}^{N}$ are the weights and bias of the convolutional filter. We then generate the video part features $P$ as follows:
\begin{equation}
p_{ij}=(\sum_{hw} A_{ihwj}f_{ihw}) / \left(HW\right).
\end{equation}
However, the attention maps may focus on background regions. To give a clear clue, we use a body part mask $M\in\mathbb{R}^{T\times H\times W \times N}$ to guide the generation of the attention maps. In detail, firstly, we use a trained segmentation model~\cite{liang2018look} to generate the part mask $M$ for input sequences. Then we resize $M$ to the same size as the attention map. Finally, $A$ and $M$ are flatted to one-dimension vectors, respectively. And a binary cross entropy loss is adopted between the flatted $A$ and corresponding flatted $M$:
\begin{equation}
\begin{aligned}
L_{p} = -\frac{1}{THWN}\sum_{b=1}^B\sum_{i=1}^{THWN} [ M_i(x_b) \log(A_i(x_b)) &\\
   + (1-M_i(x_b))\log(1-A_i(x_b)) ] &,
\end{aligned}
\label{eq10}
\end{equation}
where $B$ is the min-batch size, $x_b$ is the $b^{th}$ sequence in the batch, and $M(x_b)$ and $A(x_b)$ is the part mask and attention map of $x_b$ respectively.

\textbf{Discussion about the Generation of Spatial Relation Map.}
In the original conference paper~\cite{IANet}, the spatial relation map is generated by modeling the semantic similarity between the features of fixed space positions. That is, each position in the feature map is connected with all others and harvests semantically similar contextual information. However, there are two main limits. \textit{For one thing},~\cite{IANet} uses the semantic similarity as the correlation. In general, the features that belong to the same body part have higher semantic similarity than those belonging to different body parts. Thus,~\cite{IANet} tends to assign quite low correlations to the positions belonging to different parts, resulting in the lack of the ability to model the dependencies between different body parts. \textit{For another},~\cite{IANet} needs to generate a huge relation map to measure the semantic similarity for all position-pairs of its input. The time and space complexity are both $O(HW\times HW)$, where $H$ and $W$ denote the height and width of the input feature map respectively. Thus, when the input feature map is with high resolution, SIA~\cite{IANet} would have high computation complexity and take up huge GPU memory.

In the proposed spatial IAU (SIAU)\footnote{For image reID, STIAU is equivalent to SIAU since there are no temporal relations for input images.} for image reID, the spatial relation map is generated by modeling contextual dependencies between disjoint and distant body parts. Compared to SIA in~\cite{IANet}, SIAU has the following advantages. \textbf{1)} It can perform high-level contextual modeling between different body parts. Different from~\cite{IANet} which mainly models the dependencies within a body part, SIAU uses a sub-network to predict the correlation between different parts to capture higher-level and longer-range spatial contextual dependencies. As shown in Fig.~\ref{fig1} (b), for the two pedestrians with seemingly similar local body parts, the long-term spatial context can greatly help to clarify the local confusion thus improve the performance. \textbf{2)} It is with high computational efficiency and GPU memory friendly. Modeling relations between body parts greatly reduces the time and space complexity from $O(HW\times HW)$ to $O(N\times N)$, where $N$ ($N$$<<$$HW$) is the number of the extracted body parts in each image.

 \subsection{CIAU Module}
 \begin{figure}[t]
\centering
   \includegraphics[width=1.0\linewidth]{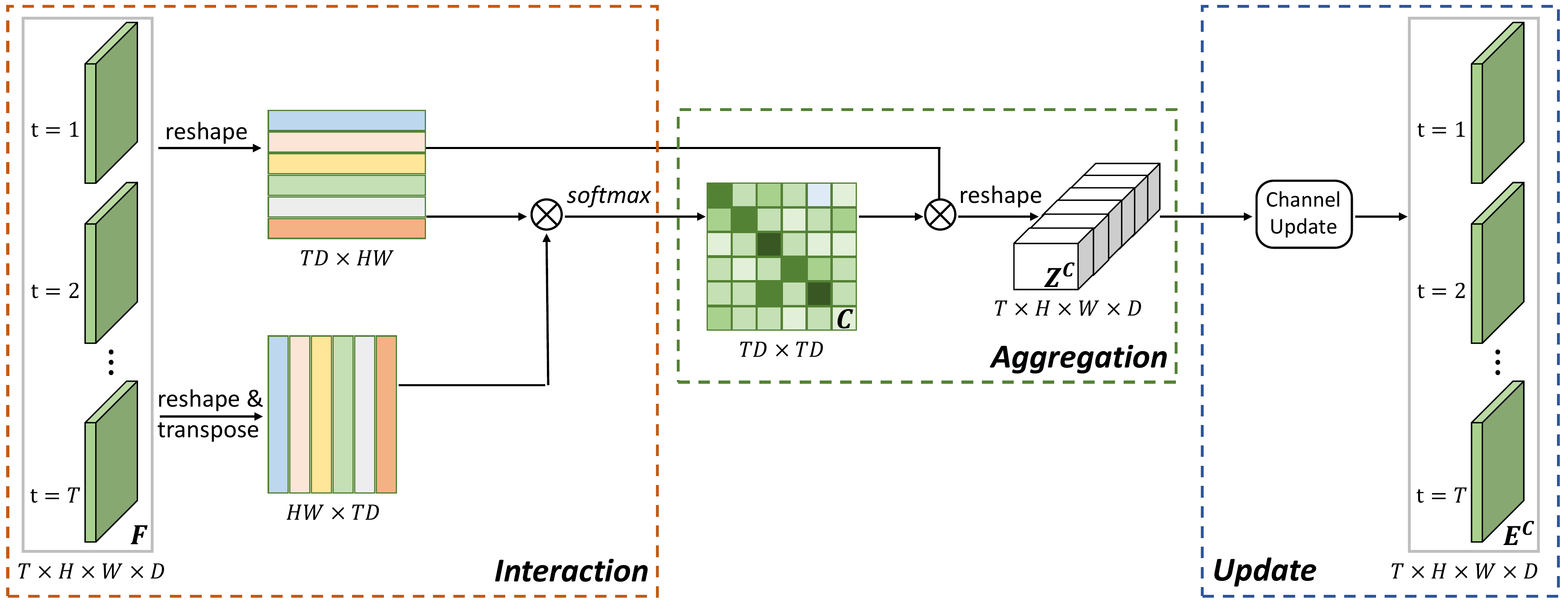}
   \caption{The architecture of CIAU module.}
\label{CIAU}
\vspace*{-1em}
\end{figure}
Existing CNN based methods typically stack multiple convolution layers to extract the features of pedestrians. With the increase of the layer number, the small-scale body parts (\textsl{e.g} shoes) easily fade away. However, these small-scale parts are very helpful to distinguish the pedestrian pairs with tiny inter-class variations. Zhang \textsl{et al.}~\cite{Occluded} have pointed out that most channel maps of high-level features show strong responses for specific parts. Inspired by their views, we build the CIAU module to aggregate semantically similar context features across all channels of a video feature map. By incorporating specific part information from other channel maps, CIAU can enhance the feature representation of that body part.

\textbf{Interaction Operation.}
As illustrated in Fig.~\ref{CIAU}, CIAU module takes a video feature map $F$ as input. In the \textit{interaction} stage, CIAU explicitly models the semantic contextual relationships between different channels of $F$ to produce a channel relation map. To this end, we first permute and reshape $F$ to $\mathbb{R}^{TD \times HW}$. Then we perform matrix multiplication between $F$ and the transpose of $F$ and normalize the results to obtain the channel relation map $C\in\mathbb{R}^{TD\times TD}$. Specifically, the semantic similarity relation between any two channels is calculated as:
\begin{equation}
\label{eq8}
C_{ij} = \frac{\exp\left(f_{i}^T f_{j}\right)}{\sum_{k=1}^{TD} \exp\left(f_{i}^T f_{k}\right)},
\end{equation}
where $f_i, f_j \in \mathbb{R}^{HW}$ denotes the features in the $i^{th}$ and $j^{th}$ channels of $F$ respectively.

\textbf{Aggregation Operation.}
Based on the channel relation map, the channel features are then aggregated in the following \textit{aggregation} operation. To be specific, a matrix multiplication between $C$ and $F$ is performed to obtain the aggregated feature map $Z^C\in\mathbb{R}^{TD\times HW}$:
\begin{equation}
Z^C = CF.
\end{equation}
$Z^C$ is then reshaped and permuted to $R^{T\times H \times W \times D}$ to maintain the input size.

\textbf{Update Operation.}
We then compute the updated channel features $E^C\in\mathbb{R}^{T\times H \times W \times D}$ based on the aggregated feature map using a \textit{channel update unit}. It is implemented by a simple convolution layer:
\begin{equation}
E^C = W_{cu} * Z^C + b_{cu},
\label{eq13}
\end{equation}
where $W_{cu}\in\mathbb{R}^{1\times 1\times D \times D}$ and $b_{cu}\in\mathbb{R}^{D}$ are the convolutional filter weights and bias. Note that the resulting feature map aggregates the contexts of different channels according to the channel relation map $C$. It is complementary to STIAU that aggregates context features of different parts according to the spatial relation map. Similar to STIAU, CIAU can adaptively adjust the input video feature map, helping to boost the feature discriminability.

\subsection{IAU block Embedding with Networks}
 \begin{figure}[t]
\centering
   \includegraphics[width=0.9\linewidth]{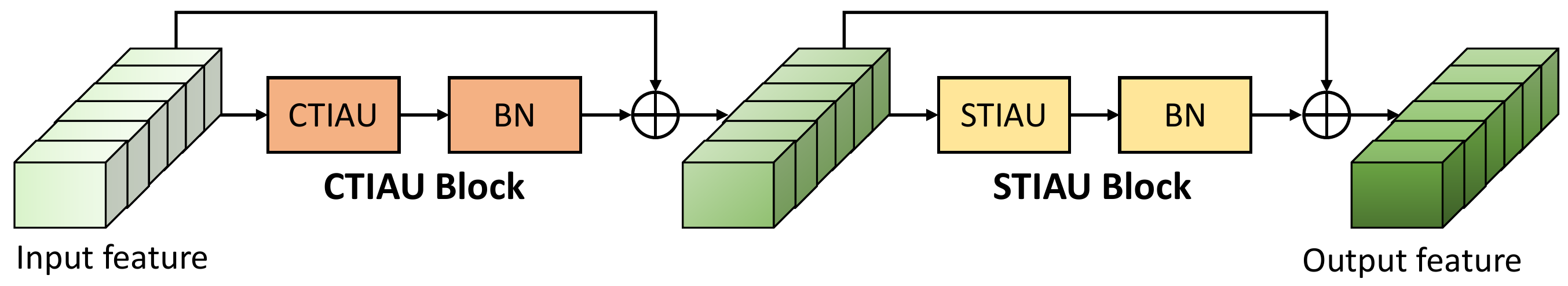}
   \caption{The architecture of IAU block.}
\label{IAU}
\end{figure}

We turn STIAU (CIAU) module into STIAU (CIAU) block which can be easily inserted into existing architectures. As shown in Fig.~\ref{IAU}, STIAU (CIAU) block is defined as:
\begin{equation}
Y=\text{BN}(E) + F.
\end{equation}
Here, $F$ is the input video feature map, $E$ is the output of STIAU or CIAU modules, and BN is a batch normalization layer~\cite{BN} to adjust the scale of $E$ to the input. A residual learning scheme ($+F$) is adopted along with the interaction-aggregation-update mechanism to facilitate the gradient flow.
Notably, before entering the BN layer, $E^S\in\mathbb{R}^{T\times D}$ is broadcasted along the spatial dimension to $\mathbb{R}^{T\times H\times W\times D}$ to be compatible with the size of $F$.

Given an input video sequence, STIAU and CIAU blocks compute complementary contextual relations. We sequentially arrange CIAU and STIAU blocks to form the IAU block, as shown in Fig.~\ref{IAU}. IAU block maintains the variable input size, thus can be inserted at any depth of networks. Considering the computational complexity, we only place it at the bottlenecks of models where the downsampling of feature maps occurs. Multiple IAU blocks located at bottlenecks of different levels can progressively boost the feature discriminability with a negligible number of parameters.

\subsection{IAUnet for Person ReID}
\begin{figure}[t]
\centering
   \includegraphics[width=1.0\linewidth]{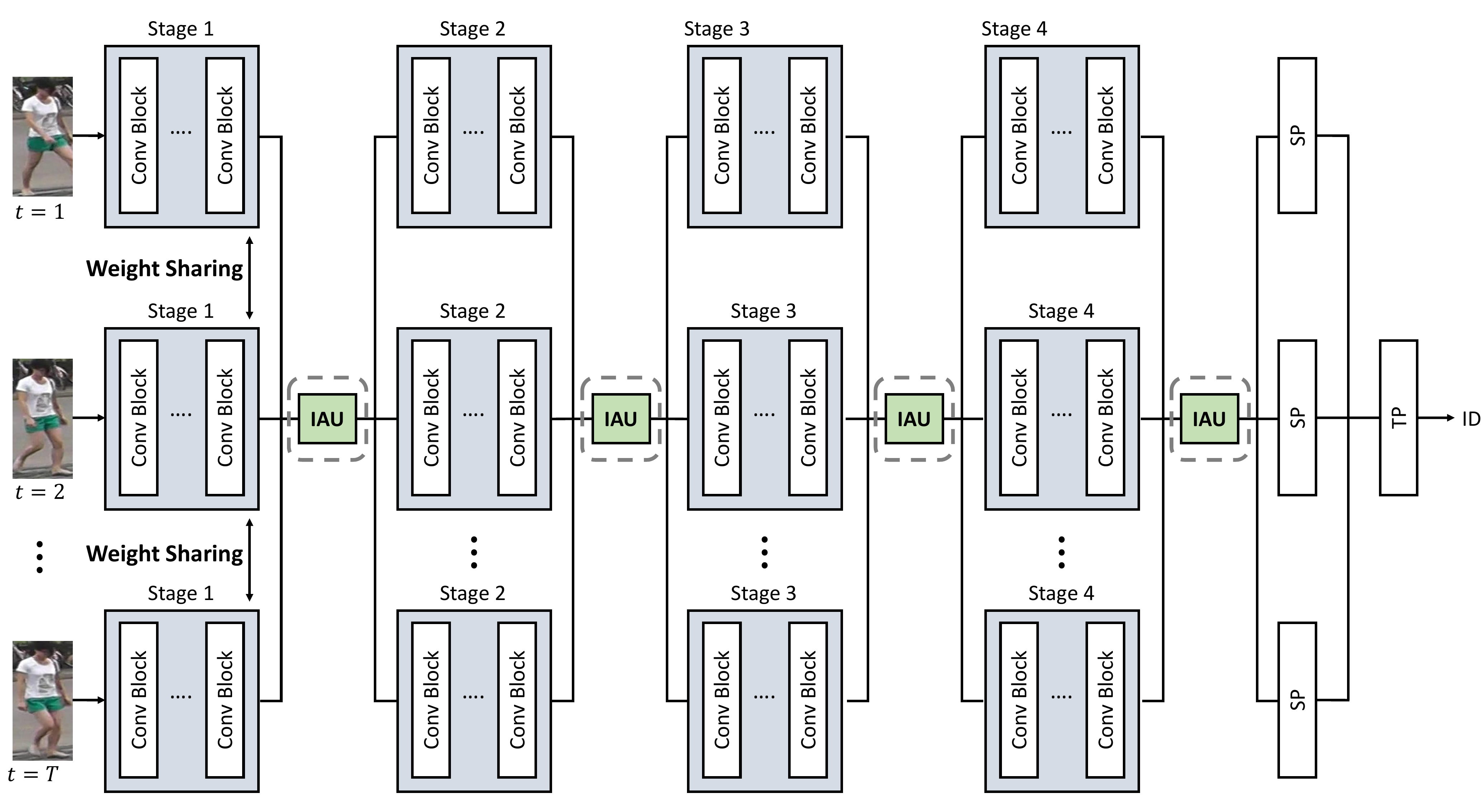}
   \caption{The architecture of IAUnet for video reID. SP and TP denote spatial pooling and temporal pooling, respectively. When the number of frames in the sequence $T$ is equal to $1$, the architecture can be used for image reID.}
\label{network}
\end{figure}
The architecture of IAUnet for person reID is illustrated in Fig.~\ref{network}. We use ResNet-50~\cite{residual} pre-trained on ImageNet~\cite{imagenet} as the backbone network, and modify the output dimension of the classification layer to the number of training identities. Besides, we remove the last spatial downsampling operation in the backbone network, which has been proven to be effective for person reID~\cite{PCB}. We denote the architecture as \textit{modified ResNet-50}. IAU blocks can be inserted into the backbone network to any stage. Different from previous works~\cite{RCN,jointly,See,snippet} that only build temporal contextual dependencies in the end, IAU blocks can capture richer temporal contextual dependencies in the earlier stages. To obtain a single feature representation for the whole sequence, a temporal average pooling layer is added in the end. Notably, the IAU block can also be used for \textbf{image person reID}, where the number of frames in the sequence is set to $1$. For image reID, IAU blocks are inserted to stage$_2$ and stage$_3$ of the backbone network. For video reID, we found that a single IAU block added to stage$_2$ gives a comparable result to multiple IAU blocks. Therefore, we only insert an IAU block to stage$_2$ of the backbone network for video reID.

\textbf{Object Function.} Following \cite{densely}, IAUnet is trained with the combination of classification and ranking. Cross Entropy Loss is used for multi-class identifies classification:
\begin{equation}
L_{cls}= -\sum_{b=1}^{B}\left[\log\left(\frac{\exp\left(p\left(y=y_b | x_b\right)\right)}{\sum_{k}\exp\left(p\left(y=k| x_b\right)\right)}\right)\right].
\end{equation}
Here $B$ is the mini-batch size, $x_b$ is the $b^{th}$ input sequence in the batch and $y_b$ is the target label of $x_b$, and $p(y|x)$ is the probability distribution of predicted label $y$ given input $x$.

Besides, we adopt a batch hard triple loss~\cite{Triplet} for correct ranking. For each sample in a batch, it only selects the hardest positive and hardest negative samples within the batch to form the triples, which is defined as:

\begin{small}
\begin{equation}
L_{tri}  = \sum_{i=1}^{C}\sum_{a=1}^{K} \left[m+\max_{s=1,\dots,K} d\left(f^i_a, f^i_s\right)-\min_{\substack{j=1,\dots,C\\ t=1,\dots,K\\ j\neq i}} d\left(f^i_a, f^j_t\right)\right]_{+}
\label{eq16}
\end{equation}
\end{small}Here, $C$ is the number of classes (person identities) of the batch, and $K$ is the number of sequences of each class. $f_j^i$ is the final extracted feature corresponding to the $j^{th}$ sequence of the $i^{th}$ person in the batch. $d$ denotes the cosine distance and $m$ is a margin hyperparameter.

Taking the spatial attention loss in STIAU into consideration, the total loss of IAU network can be denoted as:
\begin{equation}
L_{all} = L_{cls} + \lambda_{1}L_{tri} + \lambda_{2}L_{p},
\label{eq17}
\end{equation}
where $\lambda_1$ and $\lambda_2$ are the hyperparameters to balance the effects of different loss functions.

\subsection{Discussion with Other Blocks}
\label{discussion}
In this part, we give a brief discussion on the relations between the proposed IAU block and some existing blocks. The experimental comparisons can be seen in Section~\ref{related-approaches}.

\textbf{Relations to Non-local.} IAU and Non-local (NL)~\cite{non-local} can both be viewed as a graph neural network module. Compared to NL, IAU is more suitable to reID because of the following advantages: \textbf{(1)} NL is a densely-connected graph of all spatial positions of input feature maps. It requires computing a dense affinity matrix, which is computationally prohibitive for large-sized feature maps. In contrast, STIAU always has $N$ feature nodes regardless of the size of the input feature map, which is more computationally friendly. \textbf{(2)} NL captures the contextual similarity relations between the fixed geometric positions. STIAU further performs higher-level spatial-temporal contextual modeling between disjoint and distant body parts, which can alleviate the local ambiguity to better distinguish similar looking pedestrians. \textbf{(3)} NL only models the long-range contextual dependencies between spatial features. On the contrary, the proposed CIAU attempts to capture contextual knowledge between the channel features. CIAU is complementary to STIAU and conductive to highlighting important but small details or body parts.

\textbf{Relations to Squeeze-and-Excitation.} CIAU has some similarities with Squeeze-and-Excitation (SE) block~\cite{SE}. Both blocks are designed to model the contextual dependencies between channels to enhance the feature representation power. However, SE computes the channel-wise attention to selectively emphasize informative features, while it is likely to ignore the important but small parts. In contrast, CIAU aggregates the semantically similar contextual features across all channels. It can manifest the feature representations for all body parts.

\section{Experiments}
\subsection{Datasets and Settings}
We evaluate the proposed method on there image reID datasets, Market-1501~\cite{Market1501}, DukeMTMC~\cite{Duke} and MSMT17~\cite{msmt17}, two large-scale video reID datasets, MARS~\cite{mars} and DukeMTMC-VideoReID~\cite{dukereid}, an an object category classification dataset, CIFAR-100~\cite{cifar100}.

\textbf{Market-1501} is a large-scale dataset that contains $1,501$ identities. The dataset is split into two fixed parts: $12,936$ images from $751$ identities for training and $19,732$ images from $751$ identities for testing.

\textbf{DukeMTMC} is a subset of the multi-target, multi-camera pedestrian tracking dataset~\cite{multicamera}. There are $36,411$ images belonging to $1,404$ identities. It contains $16,522$ training images of $702$ identities, $2,228$ query images of the other $702$ identities, and $17,661$ gallery images.

\textbf{MTMC17} is the largest image person reID dataset, which contains $126,441$ images of $4,101$ identities. The training set contains $32,621$ images of $1,041$ identities, and the testing set contains $93,820$ images of $3,060$ identities. From the testing set, $11,659$ images are randomly selected as query images, and the others are used as gallery images.

\textbf{MARS} is the largest video reID benchmark with $1,261$ identities and $17,503$ sequences captured by $6$ cameras. It consists of $631$ identities for training and the remaining identities for testing. The bounding boxes are produced by DPM detector and GMMCP tracker, such that it provides a more challenging environment similar to real-world applications.

\textbf{DukeMTMC-VideoReID} is a subset of the tracking dataset DuKeMTMC for video reID. The dataset consists of 702 identities for training, 702 identities for testing, and 408 identities as distractors. In total there are $2,196$ videos for training and $2,636$ videos for testing.

\textbf{CIFAR100} is used to show that IAUnet can be also applied to other general recognition problems. This dataset contains 60K images of 100 classes with 600 images in each class, where 50K images are used for training and the remaining for testing.

\textbf{Evaluation Metric.}
We adopt mean Average Precision (mAP) and Cumulative Matching Characteristics (CMC) as evaluation metrics.

\textbf{Part Mask Generating.}
For each image, we first generate the part masks corresponding to four body parts: head, upper-body, lower-body, and shoes. To be specific, we first use JPPNet~\cite{liang2018look}\footnote{The code is in \url{https://github.com/Engineering-Course/LIP_JPPNet/}} pre-trained on Look into Person (LIP)~\cite{LIP} dataset to generate the part masks corresponding to $20$ semantics\footnote{Background, Hat, Hair, Glove, Sunglasses, Upper-clothes, Dress, Coat, Socks, Pants, Jumpsuits, Scarf, Skirt, Face, Right-arm, Left-arm,
Right-leg, Left-leg, Right-shoe, and Left-shoe.}. The masks of predictions for different regions are then grouped together to create $4$ coarse labels\footnote{Head, Upper-body, Lower-body, and Shoes} to guide the part division unit of the STIAU block.

\begin{table}[t]
\caption{Comparison with the state-of-the-art on Market-1501 and DukeMTMC. The methods are separated into three groups: hand-crafted methods (\textbf{H}), deep learning methods only employing global features (\textbf{G}), deep learning methods employing part features (\textbf{P}) where * denotes those requiring auxiliary part detection when testing.}
\small
\centering
\begin{tabular}{c | l |c c |c  c}
\hline
\multicolumn{2}{c|}{\multirow{2}*{Methods}}  & \multicolumn{2}{c|}{Market-1501}  &\multicolumn{2}{c}{DukeMTMC} \\
\cline{3-6}
\multicolumn{2}{c|}{ }&mAP &top-1 &mAP &top-1 \\
\hline
\multirow{3}*{\textbf{H}}&BoW+kissme~\cite{Market1501} &20.8 &44.4 &12.2 &25.1\\
&WARCA~\cite{jose2016scalable} &-- &45.2 &-- &--\\
&LOMO+XQDA~\cite{liao2015person} &-- &-- &17.0 &30.8 \\
\hline
\multirow{8}*{\textbf{G}} &SVDNet~\cite{SVdnet} &62.1 &82.3 &56.8 &76.7 \\
&GAN~\cite{Duke}  &66.1 &84.0  &47.1 &67.7 \\
&BraidNet~\cite{cascaded}  &69.5 &83.7 &59.5 &76.4 \\
&DPFL~\cite{multi-scale-representations}   &72.6 &88.6  &60.6 &79.2\\
&MLFN~\cite{Multi-Level}   &74.3 &90.0  &62.8 &81.0 \\
&KPM~\cite{KPM}  &75.3 &90.1  &63.2 &80.3 \\
&Mancs~\cite{Mancs}  &82.3 &93.1   &71.8 &84.9 \\
&Group~\cite{CRF} &81.6 &93.5  &69.5  &84.9 \\
\hline
\multirow{9}*{\textbf{P}} &Spindle*~\cite{spindle-net} &-- &76.9 &-- &-- \\
&PAR~\cite{part-aligned}  &63.4 &81.0 &-- &-- \\
&AACN*~\cite{attention-aware}    &66.9 & 85.9  &59.2 &76.8\\
&PSE*~\cite{pose-sensitive}  &69.0 &87.7   &62.0 &79.8 \\
&MGCAM~\cite{mask-guided} &74.3 &83.7 &-- &-- \\
&SPReID*~\cite{semantic} &81.3 &92.5   &70.9  &84.4\\
&RPP~\cite{PCB}  &81.6 &93.8  &69.2 & 83.3\\
&CASN~\cite{zheng2019re}  &82.8 &94.4  &73.7 &87.7\\
&IAnet~\cite{IANet} &83.1 &94.4  &73.4 &87.1\\
&DSA~\cite{zhang2019densely} & 87.6  &\textbf{95.7}  & 74.3 & 86.2 \\
\hline
&IAUnet &\textbf{88.2} &95.0  &\textbf{79.5} &\textbf{89.6}\\
\hline
\end{tabular}
\label{Tab1}
\vspace*{-0.5em}
\end{table}

\textbf{Implementation Details for Image ReID.}
For image reID, the input images are resized to $256 \times 128$. We use random flipping and random erasing~\cite{zhong2017random} with a probability of $0.5$ for data augmentation. The initial learning rate is set to $0.00035$ with a decay factor $0.1$ at every $20$ epochs. Adam~\cite{adam} optimizer is used with a mini-batch size of $64$ for $60$ epochs training. The margin of triplet loss ($m$) is set to $0.3$, and $\lambda_1$ and $\lambda_2$ (Eq,~\ref{eq17}) are set to $1$ and $0.5$ receptively.

\textbf{Implementation Details for Video ReID.}
For video reID, we randomly sample 4 frames with a stride of 8 frames from the original full-length video to form an input video sequence. The network is trained for 150 epochs in total, with an initial learning rate of $0.0003$ and reduced it with a decay rate of $0.1$ every 40 epochs. The batch size is set to $32$. Other settings and hyperparameters are the same as those in image reID.

\textbf{Implementation details for object classification.}
For object category classification, We follow the implementation details of MLFN~\cite{Multi-Level}. The initial learning rate is $0.1$ with a decay factor $0.1$ at every $100$ epochs. SGD optimization is used with a $256$ mini-batch size for $307$ epochs training. Other settings are the same as those in image reID. Notably, since there are no specific parts in the CIFIR100 images, the \textit{part division unit} of STIAU block divides equally the input feature maps into $4$ patches, and performs global average pooling on each patch to generate the corresponding part feature.

\subsection{Comparison with the State-of-the-Art Methods}
\textbf{Market-1501 and DukeMTMC.} In Tab.~\ref{Tab1}, we compare IAUnet with the state-of-the-art on Market-1501 and DukeMTMC datasets. The compared methods are categorized into three groups, \textsl{i.e.}, hand-crafted methods, deep learning methods with global features, and deep learning methods with part features. IAUnet achieves the best performance on DukeMTMC and the second best results on Market-1501. It is noted that: \textbf{(1)} The gaps between our results and those that only employ a global feature~\cite{SVdnet,cascaded,Duke,multi-scale,KPM,Mancs,Cross-view} are significant: about $7\%$ mAP improvement. The significant improvements demonstrate that it is effective to employ the spatial contextual information for reID. \textbf{(2)} Some part-related methods incorporate an external part detection network~\cite{spindle-net,semantic,pose-sensitive,zhang2019densely} into the reID model, which makes the reID model too complicated and time consuming. IAUnet puts much fewer overheads with much better performance on DukeMTMC: about $5\%$ mAP improvement. We argue that the improvement is due to the alleviation of local confusion by modeling the global contextual relations between different body parts. \textbf{(3)} Other attention-centric methods~\cite{part-aligned,mask-guided,attention-aware,PCB,IANet} use lightweight attention subnet to learn discriminative body parts. IAUnet outperforms these methods with an improvement up to $6\%$ on mAP. The superiority of IAUnet over the attention-centric methods further verifies the effectiveness of modeling spatial contextual dependencies among different parts.

\textbf{MSMT17.}
\begin{table}[t]
\caption{Comparison with the state-of-the-art on MSMT17.}
\small
\centering
\begin{tabular}{l | c c c c}
\hline
\multirow{2}*{Methods} &\multicolumn{4}{c}{MSMT17} \\
\cline{2-5}
&mAP &top-1 &top-5 &top-10 \\
\hline
GoogleNet~\cite{going} &23.0 &47.6 &65.0 &71.8 \\
Pose-driven~\cite{pose-driven} &29.7 &58.0 &73.6 &79.4\\
GLAD~\cite{Glad}&34.0  &61.4 &76.8 &81.6 \\
IANet~\cite{IANet} &46.8 &75.5 &85.5 &88.7 \\
\hline
IAUnet &\textbf{59.9} &\textbf{82.0} &\textbf{90.5} &\textbf{93.1}  \\
\hline
\end{tabular}
\label{tab2}
\end{table}
\begin{table}[t]
\small
\centering
\caption{Comparison with related methods on MARS. The methods are separated into two groups: deep learning methods only employing global video features (\textbf{G}), deep learning methods employing temporal attention module (\textbf{A}).}
\vspace*{-0.5em}
\label{mars}
\begin{center}
\begin{tabular}{c | l | c c c c}
\hline
\multicolumn{2}{c|}{\multirow{2}*{Methods}} &\multicolumn{4}{c}{MARS} \\
\cline{3-6}
\multicolumn{2}{c|}{ } &mAP &top-1 &top-5 &top-10  \\
\hline
\multirow{5}*{\textbf{G}}&Mars \cite{mars}  &49.3 &68.3 &82.6 &89.4  \\
&Latent Parts \cite{context-aware}    &56.1 &71.8 &86.6 &93.0  \\
&K-reciprocal  \cite{re-rank}    &68.5 &73.9 &--    &--     \\
&EUG \cite{dukereid} & 67.4  &80.8 &92.1 &96.1 \\
&TriNet       \cite{Triplet}     &67.7 &79.8 &91.4 &--   \\
\hline
\multirow{8}*{\textbf{A}}&SeeForest  \cite{See}   &50.7  &70.6 &90.0 &97.6 \\
&Seq-Decision  \cite{sequence-decision}  &-- &71.2 &85.7 &91.8  \\
&QAN          \cite{QAN}   &51.7  &73.7 &84.9 &91.6 \\
&STAN         \cite{diversity}  &65.8  &82.3 &--    &--     \\
&RQEN        \cite{RQAN}   &71.7   &77.8 &88.8 &94.3 \\
&Snippet      \cite{snippet}  &76.1 &86.3 &94.7 &98.2  \\
&TAFD \cite{zhao2019attribute}  &78.2 &87.0 &95.4 &98.7 \\
&VRSTC \cite{VRSTC} &82.3 &88.5 &96.5 &97.4  \\
\hline
&IAUnet   &\textbf{85.0}    &\textbf{90.2} &\textbf{96.6}  &\textbf{98.3} \\
\hline
\end{tabular}
\end{center}
\label{tab3}
\vspace*{-0.5em}
\end{table}
We further evaluate the proposed method on a recent large scale dataset, namely MSMT17. As shown in Tab.~\ref{tab2}, the proposed method significantly outperforms existing works~\cite{going,pose-driven,Glad} with a top-1 accuracy of $20.6\%$ and mAP of $25.9\%$. IANet models the \textbf{intra-parts} contextual dependencies to adaptively locate the body parts. IAUnet significantly outperforms it with an improvement of up to $13.1\%$ on mAP, which demonstrates the superiority of \textbf{inter-parts} contextual dependencies modeling.

\textbf{MARS.}
Tab.~\ref{tab3} reports the performance of IAUnet and current state-of-the-art on MARS. The proposed method outperforms the best existing methods. \textbf{(1)} The works that only employ global video features (\textbf{G})~\cite{mars,context-aware,re-rank,Triplet,dukereid} treat each frame of a video equally, resulting in the corruption of the video representation by mis-detected frames. IAUnet surpasses these works by up to $10\%$ and $18\%$ on top-1 accuracy and mAP respectively. \textbf{(2)} Other attention-based works (\textbf{A})~\cite{See,sequence-decision,QAN,RQAN,diversity,snippet,zhao2019attribute} leverage a temporal attention network to select the most discriminative frames, resulting in the loss of spatial-temporal information of the video. IAUnet outperforms these works up to $3\%$. The improvement can be attributed to the feature enhancement by capturing richer spatial-temporal contextual dependencies in IAU blocks.

\textbf{DukeMTMC-VideoReID.}
\begin{table}[!t]
\small
\centering
\caption{Comparisons on DukeMTMC-VideoReID.}
\vspace*{-0.5em}
\label{duke}
\begin{center}
\begin{tabular}{l | c c c c}
\hline
\multirow{2}*{Methods} &\multicolumn{4}{c}{DukeMTMC-VideoReID} \\
\cline{2-5}
&mAP &top-1 &top-5 &top-10  \\
\hline
EUG \cite{dukereid} &78.3 &83.6 &94.6 &97.6  \\
VRSTC \cite{VRSTC} &93.8 &95.0 &99.1 &99.4 \\
\hline
IAUnet &\textbf{96.1} &\textbf{96.9} &\textbf{99.5} &\textbf{99.8}  \\
\hline
\end{tabular}
\end{center}
\label{tab4}
\end{table}
As shown in Tab.~\ref{tab4}, the proposed method outperforms the current best result of $1.9\%$ and $2.3\%$ in top-1 accuracy and mAP respectively on DukeMTMC-VideoReID. VRSTC~\cite{VRSTC} uses a completion network to recover the appearance of occluded regions as a pre-processing. It is orthogonal to IAUnet and can be easily combined to further improve the performance.

\begin{table}[t]
\caption{Object classification results on CIFAR-100 dataset. * indicates results reported by MLFN \cite{Multi-Level}.}
\small
\centering
\begin{tabular}{l |c}
\hline
\multirow{2}*{Methods} &\multicolumn{1}{c}{CIFAR-100} \\
\cline{2-2}
&Error Rates (\%) \\
\hline
ResNet-50* & 30.21  \\
ResNeXt-50* & 29.03 \\
DualNet & 27.57 \\
MLFN* & 27.21 \\
\hline
IAUnet & \textbf{20.30} \\
\hline
\end{tabular}
\label{tab5}
\end{table}

\subsection{Object Categorization Results}
In this part, we evaluate IAUnet on a more general object classification task by experimenting on CIFAR-100. Tab.~\ref{tab5} compares IAUnet with ResNet-50~\cite{identity}, ResNeXt-50~\cite{Aggregated}, DualNet~\cite{Dualnet} and MLFN~\cite{Multi-Level}. ResNet-50~\cite{identity}, ResNeXt-50~\cite{Aggregated}, and MLFN~\cite{Multi-Level} have similar depth and model sizes to IAUnet. The improved result over ResNet-50 shows that IAU blocks bring obvious benefit. IAUnet also outperforms MLFN~\cite{Multi-Level} that fuses multi-scale features. Besides, IAUnet beats DualNet~\cite{Dualnet} that fuses two complementary ResNet branches in an ensemble and has double model size. The consistent improvements suggest that IAU blocks can be easily generalized to general recognition problems.

\subsection{Comparison with Related Approaches}
\label{related-approaches}

\begin{table}[t]
\caption{Comparison to Non-local (NL) and Squeeze-and-Excitation (SE) on both image and video reID.}
\centering
\small
\subfloat[Image reID dataset Market-1501.]{
\begin{tabular}{l |c c |c  c}
\hline
\multirow{2}*{Models}  & \multicolumn{4}{c}{Market-1501} \\
\cline{2-5}
&Param. &GFLOPs &mAP &top-1 \\
\hline
baseline &23.5M &4.06 &84.5 &93.4  \\
\hline
NL~\cite{non-local} &26.1M &4.75 &86.2 &94.2 \\
STIAU &26.1M &4.07 &\textbf{87.3} &\textbf{94.8} \\
\hline
SE~\cite{SE} &23.7M &4.06 &85.4 &93.9 \\
CIAU &24.8M &4.86 &\textbf{86.8} &\textbf{94.2} \\
\hline
IAUnet &27.4M &4.87 &\textbf{88.2} &\textbf{95.0}\\
\hline
\end{tabular}
\label{taba}
}

\subfloat[Video reID dataset MARS.]{
\begin{tabular}{l |c c |c  c}
\hline
\multirow{2}*{Models}  & \multicolumn{4}{c}{MARS} \\
\cline{2-5}
&Param. &GFLOPs &mAP &top-1 \\
\hline
baseline  &23.5M &16.24 &83.5 &88.2\\
\hline
NL~\cite{non-local}  &24.0M &19.46 &84.0 &88.8\\
STIAU &24.0M &16.25 &\textbf{84.9} &\textbf{89.6}\\
\hline
SE~\cite{SE}  &23.5M &16.24 &83.5 &88.7\\
CIAU &23.7M &21.07 &\textbf{84.5} &\textbf{89.1}\\
\hline
IAUnet &24.3M &21.08 &\textbf{85.0} &\textbf{90.2}\\
\hline
\end{tabular}
\label{tabb}
}
\vspace*{-1em}
\label{tab6}
\end{table}

In this section, we present the experimental results compared to non-local (NL)~\cite{non-local} and squeeze-and-excitation (SE)~\cite{SE} blocks mentioned in Section~\ref{discussion}. The results are summarized in Tab.~\ref{tab6}. We adopt the \textit{modified ResNet-50} model as the baseline and replace the IAU blocks in IAUnet with NL, STIAU, SE, and CIAU blocks representatively.

\textbf{Comparison STIAU with Non-Local.} As shown in Tab.~\ref{tab6}, compared with the non-local method, STIAU blocks show higher accuracy under the same model size and less computation budge.
We argue that it is hard for non-local to directly reason the contextual relations between different body parts. Instead, STIAU blocks are designed to explicitly capture the contextual dependencies between spatially distant parts regardless of their shapes. It can provide complementary features that can not be easily captured by stacking convolution layers and non-local blocks.

\textbf{Comparison CIAU with Squeeze-and-Excitation.} As shown in Tab.~\ref{tab6}, we can observe that CIAU achieves better accuracy than SE. We use the default hyperparameters in~\cite{SE} for SE which leads to marginal improvements. CIAU significantly outperforms SE by $1.4\%$ and $2\%$ mAP on Market-1501 and MARS respectively. The results indicate that CIAU can better model the contextual interdependencies between channels. The significant improvements also demonstrate that it is more efficient to enhance feature representation power by aggregating similar channel features than multiplying them by constants. Furthermore, by combining CIAU with STIAU, the performance can be further lifted for both image and video reID tasks.

\subsection{Ablation Study}

To investigate the effectiveness of each component in the IAU block, we conduct a series of ablation studies on two image reID datasets: Market-1501 and DukeMTMC, and two video reID datasets: MARS and DukeMTMC-VideoReID. Tab~\ref{tab7} and Tab.~\ref{tab8} summarize the comparison results for different settings. We adopt \textit{modified ResNet-50} as the baseline. If there is no special explanation, the proposed blocks are inserted into the last residual block of the stage$_2$ layer of \textit{modified ResNet-50}.

\textbf{STIAU blocks.} As shown in Tab.~\ref{tab7} and Tab.~\ref{tab8}, STIAU blocks consistently improve the performance remarkably. For image reID, the STIAU block brings about $2\%$ mAP improvements over the baseline. We further compare STIAU to SIA block~\cite{IANet}. As shown in Tab.~\ref{tab7}, STIAU outperforms SIA by about $1\%$ mAP and top-1 accuracy, indicating that STIAU can better models the spatial relations. For video reID, we study the effect of STIAU blocks applied along spatial, temporal, and spatial-temporal dimensions. For example, in the spatial-only version, the contextual dependencies only happen within the same frame: \textit{i.e.}, $R$ (in Eq.~\ref{eq3}) is simply set to $S$ (in Eq.~\ref{eq1}). Accordingly, the temporal-only version sets $R$ to $T$ (in Eq.~\ref{eq2}). Tab.~\ref{tab8} shows that both the spatial-only and temporal-only versions improve over the baseline, and the performance can be further lifted when the spatial and temporal contextual dependencies are integrated into the STIAU block.

\begin{table}[t]
\caption{Performance comparisons of baseline and proposed schemes on image reID task.}
\small
\centering
\begin{tabular}{ l |c c |c  c}
\hline
\multirow{2}*{Methods}  & \multicolumn{2}{c|}{Market-1501}  &\multicolumn{2}{c}{DukeMTMC} \\
\cline{2-5}
&mAP &top-1 &mAP &top-1 \\
\hline
Baseline &84.5 &93.4 &76.2 &87.8\\
\hline
SIA~\cite{IANet} & 85.3 & 93.8 &  77.1 & 88.2\\
STIAU &\textbf{86.5} &\textbf{94.6} &\textbf{78.2} &\textbf{89.1}\\
\hline
CIAU &86.3 &94.4 &78.1 &88.9\\
IAU &\textbf{86.7} &\textbf{94.8} &\textbf{78.9} &\textbf{89.2}\\
\hline
IAU (stage$_1$) &85.6 &93.8 &77.3 &88.2\\
IAU (stage$_2$) &86.7 &\textbf{94.8} &78.9 &89.2\\
IAU (stage$_3$) &\textbf{86.8} &94.3 &\textbf{79.3} &\textbf{89.3}\\
IAU (stage$_4$) &85.1 &93.8 &76.7 &87.6\\
\hline
IAU-w/o-$L_p$ (stage$_{23}$) &87.3 &94.7 &78.5 &88.9\\
IAU (stage$_{23}$) &\textbf{88.2} &\textbf{95.0} &\textbf{79.5} &\textbf{89.6}\\
\hline
\end{tabular}
\label{tab7}
\vspace*{-0.5em}
\end{table}
\begin{table}[t]
\caption{Performance comparisons of baseline and proposed schemes on video reID task.}
\small
\centering
\begin{tabular}{ l |c c |c  c}
\hline
\multirow{2}*{Methods}  & \multicolumn{2}{c|}{MARS}  &\multicolumn{2}{c}{Duke-Video} \\
\cline{2-5}
&mAP &top-1 &mAP &top-1 \\
\hline
baseline &83.5 &88.2 &94.5 &95.0\\
\hline
STIAU-spatial-only  &84.6 &88.9 &95.3 &96.0\\
STIAU-temporal-only &84.6 &89.1 &95.0 &95.8\\
STIAU &\textbf{84.9} &\textbf{89.6} &\textbf{95.7} &\textbf{96.2}\\
\hline
CIAU &84.5 &89.1 &95.6 &96.0\\
IAU & \textbf{85.0}&\textbf{90.2} &\textbf{96.1} &\textbf{96.9}\\
\hline
IAU-w/o-$L_p$ (stage$_{2}$) &84.5 &88.2 &95.3 &96.2 \\
IAU (stage$_2$) &\textbf{85.0} &\textbf{90.2} & \textbf{96.1} &\textbf{96.9}\\
\hline
IAU (stage$_3$) &84.5 &89.1 &96.0 &96.8\\
IAU (stage$_{23}$) &\textbf{85.3} &90.0 &95.9 &96.3\\
\hline
\end{tabular}
\label{tab8}
\vspace*{-0.5em}
\end{table}

\textbf{CIAU blocks.} We further assess the effectiveness of the CIAU block by adding it to the baseline. CIAU individually outperforms the baseline by about $2\%$ and $1\%$ in terms of mAP and top-1 accuracy on image datasets and video datasets respectively. The improvements indicate that it is effective to enhance feature representation power by aggregating similar features along the channel dimension. When we integrate STIAU and CIAU blocks to the IAU block, the performance can be further improved by about $1\%$ on mAP and top-1 accuracy. We argue that the STIAU and CIAU capture the complementary contextual dependencies, spatial, and channel. This leads to each block can provide some complementary features that cannot be easily captured by another block.

\begin{table}[t]
\caption{{Comparisons of different relation calculation strategies on Market-1501 and MARS.}}
\small
\centering
\begin{tabular}{ l |c c |c  c}
\hline
\multirow{2}*{Methods}  & \multicolumn{2}{c|}{Market-1501}  &\multicolumn{2}{c}{MARS} \\
\cline{2-5}
&mAP &top-1 &mAP &top-1 \\
\hline
STIAU-L2 &84.7 &93.5 &83.4 &88.1\\
STIAU &\textbf{86.5} &\textbf{94.6} &\textbf{84.9} &\textbf{89.6}\\
\hline
CIAU-L2 &85.7 &94.1 &84.0 &88.8\\
CIAU &\textbf{86.3} &\textbf{94.4} &\textbf{84.5} &\textbf{89.1}\\
\hline
\end{tabular}
\label{r-tab1}
\vspace*{-0.5em}
\end{table}

\textbf{Relation calculation strategy.} We exploit another relation calculation strategy, \textsl{i.e.}, L2 distance, for STIAU and CIAU blocks. Firstly, Tab.~\ref{r-tab1} compares STIAU and STIAU-L2, where STIAU and STIAU-L2 use a sub-network and L2 distance respectively to calculate the relations between different parts. We can see that STIAU significantly outperforms STIAU-L2. We argue that the distance metric only models the semantic similarity dependencies, while the sub-network can model higher-level dependencies. For instance, since the head parts typically have highly accurate pedestrian characteristics, the sub-network can learn to assign high relations between the head part and other body parts. Then other body parts can integrate the features of the head part to improve their discrimination. Secondly, Tab.~\ref{r-tab1} also compares CIAU and CIAU-L2 where CIAU and CIAU-L2 use dot-product and L2 distance respectively to calculate the relations for channel-pairs. The dot-produce and L2 distance both belong to the distance metric, but dot product is more implementation-friendly in deep learning platforms. We can see that CIAU performs slightly better than CIAU-L2.

\begin{table}[t]
\caption{{Combining methods of STIAU and CIAU blocks on Market-1501 and MARS.}}
\small
\centering
\begin{tabular}{ l |c c |c  c}
\hline
\multirow{2}*{Methods}  & \multicolumn{2}{c|}{Market-1501}  &\multicolumn{2}{c}{MARS} \\
\cline{2-5}
&mAP &top-1 &mAP &top-1 \\
\hline
STIAU+CIAU &85.6 &94.3 & 84.5&88.2 \\
STIAU-CIAU &85.9 &94.4 &84.4 &88.6 \\
CIAU-STIAU &\textbf{86.7} &\textbf{94.8} &\textbf{85.0} &\textbf{90.2}\\
\hline
\end{tabular}
\label{r-tab2}
\vspace*{-0.5em}
\end{table}

\textbf{Arrangement of STIAU and CIAU blocks.} We compare three different ways of arranging STIAU and CIAU blocks: sequential STIAU and CIAU blocks (STIAU-CIAU), sequential CIAU and STIAU blocks (CIAU-STIAU), and parallel use of both blocks (STIAU+CIAU). As each block has different functions, the combination mode and order may affect the overall performance. Tab.~\ref{r-tab2} summarizes the experimental results on different arranging methods, where CIAU-STIAU produces the best results on both image and video reID tasks. We argue that STIAU and CIAU blocks compute complementary contextual relations, where STIAU blocks focus on ``spatial-temporal'' and ``different parts'' modeling while CIAU blocks focus on ``channel'' and ``same part'' modeling. The CIAU blocks, which can enhance the representation of the individual body part, is conducive to the relationship modeling between different parts by STIAU blocks. Thus the sequential CIAU-STIAU can achieve the best performance.

\textbf{Efficient positions to place IAU blocks.} Tab.~\ref{tab7} compares a single IAU block added to the different stages of ResNet50. The block is added to right before the last residual block of a stage. The improvements of an IAU block in stage$_2$ and stage$_3$ are similar but smaller in stage$_1$ and stage$_4$. One possible explanation is that stage$_1$ has a big spatial size that is not very expressive and sufficient to provide precise semantic information. Besides, the visual concepts in stage$_4$ tend to be too abstract thus it is difficult to aggregate context features in this stage. Therefore, we only consider adding IAU blocks to stage$_2$ and stage$_3$ layers.

\begin{table}[t]
\caption{Performance gain by adding IAU blocks on different networks on Market-1501 and MARS.}
\small
\centering
\begin{tabular}{ l| l |c c |c  c}
\hline
\multirow{2}*{Backbone} & \multirow{2}*{Method} & \multicolumn{2}{c|}{Market-1501}  &\multicolumn{2}{c}{MARS} \\
\cline{3-6}
&&mAP &top-1 &mAP &top-1 \\
\hline
\multirow{2}*{ResNet18} & baseline &72.0 &89.1 &74.9 &83.7\\
&ResNet18-IAU &\textbf{74.9}&\textbf{89.8} &\textbf{76.4} &\textbf{84.8} \\
\hline
\multirow{2}*{ResNet34} & baseline &74.8 &89.5 &78.4 &85.9\\
&ResNet34-IAU &\textbf{78.5} &\textbf{91.0} &\textbf{79.7} &\textbf{86.9}\\
\hline
\multirow{2}*{Inception} & baseline &72.2 &88.1 &72.7 &82.2\\
&Inception-IAU &\textbf{75.2} &\textbf{89.7} &\textbf{74.1} &\textbf{83.6}\\
\hline
\end{tabular}
\label{tab9}
\vspace*{-0.5em}
\end{table}

\textbf{Going deeper with IAU blocks.} Tab.~\ref{tab7} and Tab.~\ref{tab8} also show the results of more IAU blocks. For image reID, IAU blocks can consistently lift the accuracy when more blocks are added. In particular, The model with IAU blocks added to stage$_2$ and stage$_3$ (IAU-stage$_{23}$) improves the model with one IAU block added to stage$_2$ or stage$_3$ (IAU-stage$_2$ or IAU-stage$_3$) by about $1.5$ mAP on Market-1501. We argue that multiple IAU blocks can perform hierarchical communication, where each block can provide some complementary relations which can not be easily captured by other blocks. For video reID, we find that adding two IAU blocks does not give significant gain as shown in the last three rows of Tab.~\ref{tab8}. Therefore, we only add a single IAU block to stage$_2$ of the backbone network for video reID.

\textbf{Effect of the spatial attention constrain $L_p$.}
To evaluate the contribution of the proposed spatial attention constrain $L_p$, we train IAUnet and report the results without spatial attention constrain $L_p$ (\textbf{IAU-w/o-$L_p$}). Experimental results are presented in Tab.~\ref{tab7} and Tab.~\ref{tab8}. We can observe that the results of IAUnet consistently outperform that of IAU-w/o-$L_p$ on both image and video reID benchmarks. This confirms the effectiveness of using spatial attention constrain in IAUnet. We argue that without the spatial attention constrain, the learned multiple attention maps tend to be disorganized and focus on the same regions, as shown in Fig.~\ref{attention} (b). Therefore, it is difficult for IAU-w/o-$L_p$ to establish global contextual dependencies between different body parts, resulting in performance degradation.

\textbf{Effectiveness of IAU blocks across different backbone.}
We then investigate the generality of IAU blocks on different CNNs. We firstly investigate the effect of combining IAU blocks with ResNet18~\cite{residual} and ResNet34~\cite{residual}. The results are summarized in Tab.~\ref{tab9}, where all baseline results are reproduced by ourselves using the same training schema for a fair comparison. We can observe the significant performance improvement induced by IAU blocks. In particular, ResNet18-IAU has a mAP of $89.8\%$ on Market-1501, which is superior to both its direct counter part ResNet18 ($79.1\%$) as well as the deeper ResNet34 ($89.5\%$). We further exam the effect of IAU blocks on the \textit{non-residual} network by experimenting with Inception architecture~\cite{going}. We can observe the same phenomena that emerged in the residual architectures. Overall, these experiments demonstrate that IAU blocks can consistently boost the accuracy of a wide range of architectures on both image and video reID tasks.

\begin{table}[t]
\caption{Performance gain by adding IAU blocks on the existing reID framework on Market-1501 and DukeMTMC.}
\small
\centering
\begin{tabular}{ l |c c |c  c}
\hline
\multirow{2}*{Methods}  & \multicolumn{2}{c|}{Market-1501}  &\multicolumn{2}{c}{DukeMTMC} \\
\cline{2-5}
&mAP &top-1 &mAP &top-1 \\
\hline
PCB~\cite{PCB} &78.4 &91.9 &66.1 &81.8\\
PCB-IAU &\textbf{79.9} &\textbf{92.7} &\textbf{70.4} &\textbf{83.8}\\
\hline
CAMA~\cite{CAMA} &84.5 &94.7 &72.9 &85.8\\
CAMA-IAU &\textbf{87.2} &\textbf{95.2} &\textbf{78.7} &\textbf{89.4}\\
\hline
\end{tabular}
\label{r-tab3}
\vspace*{-0.5em}
\end{table}

\textbf{Effectiveness of IAU blocks across existing reID methods.}
Finally, we try another two person reID frameworks, PCB~\cite{PCB} and CAMA~\cite{CAMA}, to further verify the generality of the proposed IAU block. The results are summarized in Tab.~\ref{r-tab3}, where the IAU blocks are added to $\text{stage}_2$ and $\text{stage}_3$ of the backbone of PCB and CAMA to form PCB-IAU and CAMA-IAU networks respectively. We can observe the significant performance improvement induced by IAU blocks, showing the generality of IAU blocks.

\subsection{Visualizing STIAU Block}
\begin{figure}[t]
\centering
   \includegraphics[width=1.0\linewidth]{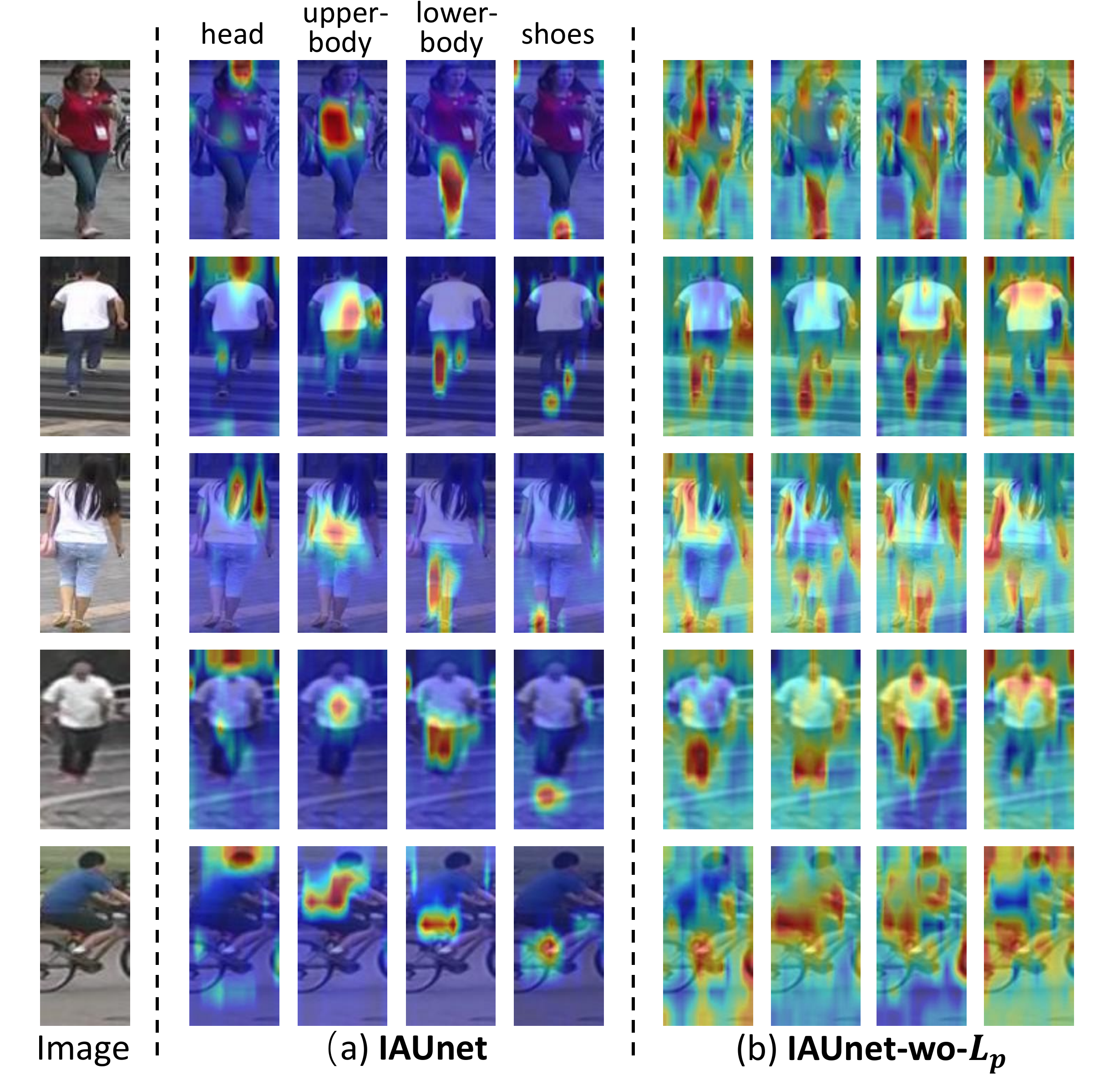}
   \caption{\textbf{Learned Spatial Attention Maps.} Example images and corresponding receptive fields for part-specific attention maps when $N=4$. (a) Visualization of the IAUnet. (b) Visualization of the IAUnet trained without spatial attention constrain $L_p$.}
\label{attention}
\vspace*{-1.5em}
\end{figure}
In this subsection, we visualize the learned spatial attention maps of the part division unit in the STIAU block. Fig.~\ref{attention} (a) shows the four spatial attention maps generated by IAUnet for five images. As expected, different spatial attention maps attempt to focus on different local body parts, \textit{i.e.}, head, upper-body, lower-body, and shoes. It is noteworthy that the spatial attention maps can adaptively localize the body parts under various challenging situations, such as small scale (the second row), motion blur (the fourth row) even dramatic changes in pose (the last row). Each attention map is used for a weighted average pooling over the whole image, producing a single body-part feature for globally spatial-temporal contextual dependencies modeling. Without the part features, it is really difficult for convolution operations to directly model contextual dependencies between such patterns that might be spatially distant or ill-shaped.

\subsection{Visualizing Relation Maps of STIAU and CIAU blocks}
\begin{figure}[t]
\centering
   \includegraphics[width=0.75\linewidth]{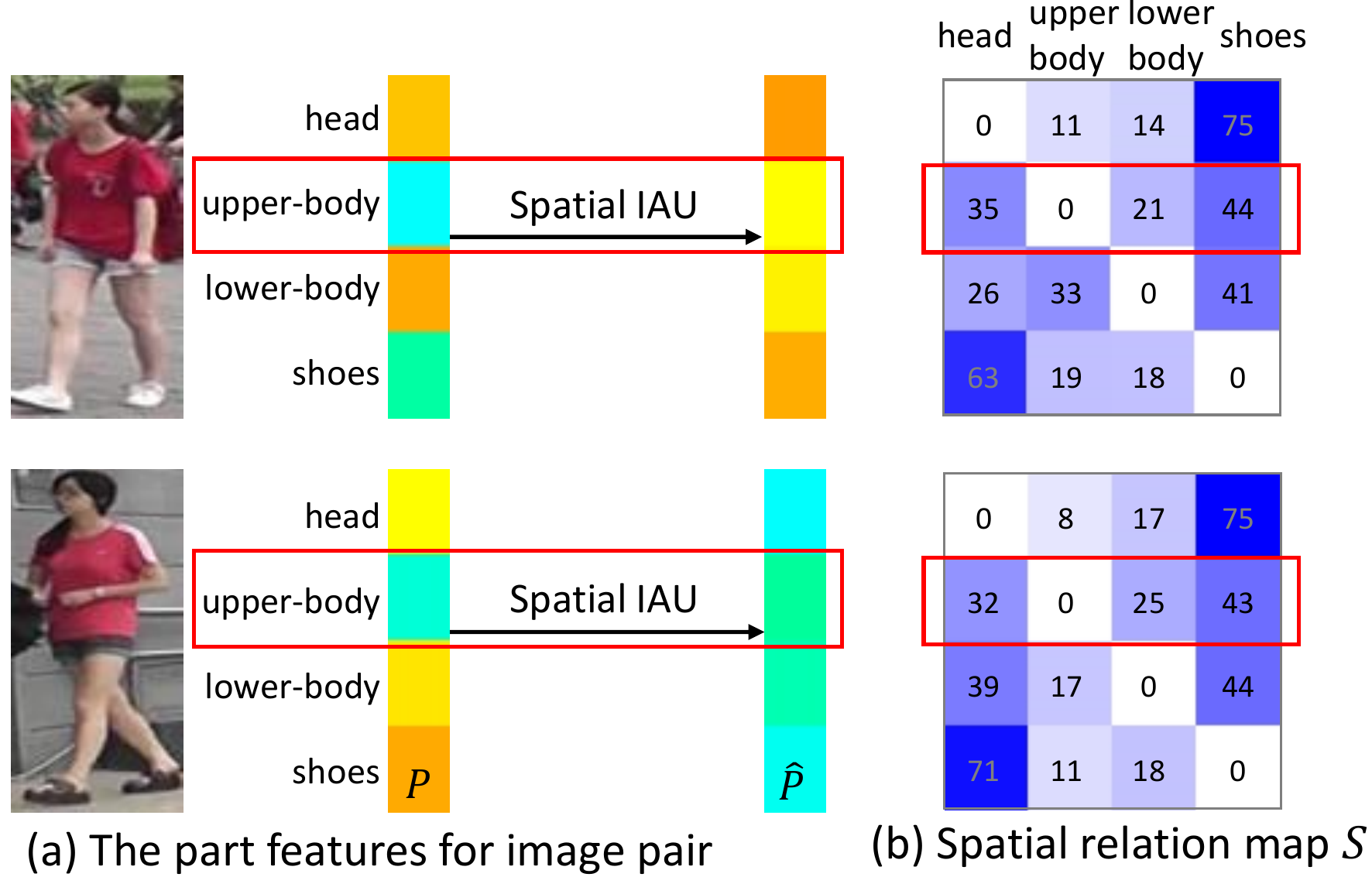}
   \caption{ \textbf{Learned Spatial Relation Map.} (a) Visualization of the initial part features $P$ and updated part features $\hat{P}$ by spatial IAU for input image pair. The dimensionality of $P$ and $\hat{P}$ is reduced to $N\times1$ ($N=4$) by PCA for visualization. (b) the spatial relation maps $S$ with size $N\times N$.}
\label{r-SIAU}
\vspace*{-1.5em}
\end{figure}

In this part, we visualize the learned spatial, temporal, and channel relation maps respectively. Fig.~\ref{r-SIAU} visualizes the initial part features ($P$ in Fig.~\ref{STIAU}), the updated part features by spatial IAU ($\hat{P}$ in Fig.~\ref{STIAU}), and the spatial attention maps ($S$ in Fig.~\ref{STIAU}) of spatial IAU. It is clear that, for the two persons with similar upper clothes, the initial upper-body features $P$ are difficult to distinguish between the two persons. The spatial relation map $S$ stores global spatial relations. As shown in Fig.~\ref{r-SIAU}, for the upper body part, $S$ assigns larger correlation values to the shoes and head parts that are highly discriminative for the input two persons. Therefore, with the feature propagation through $S$, the upper body features can be updated to distinguish the two persons, as shown in Fig.~\ref{r-SIAU}. In addition, we can observe that the head and shoe parts tend to present larger values in $S$. We argue that the head and shoe parts usually have more accurate pedestrian characteristics than body clothes. Thus, spatial IAU learns to assign high correlations between the head/shoe parts and other body parts, so that the other body parts can integrate the features of the head and shoe parts to improve their discrimination.
\begin{figure}[t]
\centering
   \includegraphics[width=0.85\linewidth]{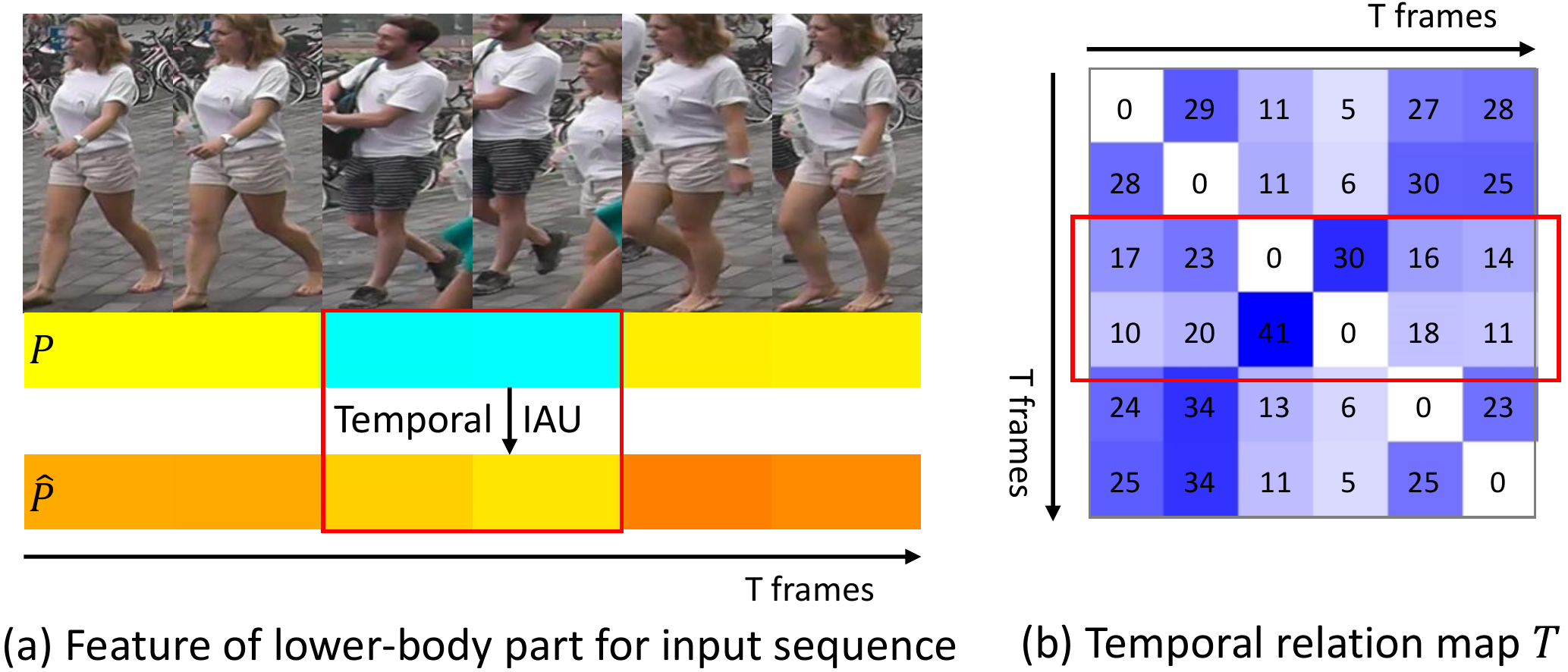}
   \caption{ \textbf{Learned Temporal Relation Map.} (a) Visualization of the initial part features $P$ and updated part features $\hat{P}$ by temporal IAU for the input sequence. The dimensionality of $P$ and $\hat{P}$ is reduced to $T\times1$ ($T=6$) by PCA for visualization. (b) the temporal relation maps $T$ with size $T\times T$.}
\label{r-TIAU}
\vspace*{-1.5em}
\end{figure}

Fig.~\ref{r-TIAU} visualizes the temporal relation map ($T$ in Fig.~\ref{STIAU}) for the lower-body part of the input sequence. It also visualizes the initial lower-body part features ($P$ in Fig.~\ref{STIAU}) and updated lower-body part features ($\hat{P}$ in Fig.~\ref{STIAU}) by temporal IAU module. It is clear that, the detection errors affect the initial part feature, i.e., the feature substantially changes as misdetection happens. The temporal relation map stores the global temporal contextual relations. As shown in Fig.~\ref{r-TIAU}, for the mis-detected frames, $T$ assigns more than $50\%$ weights to the good frames. Therefore, with the feature propagation through $T$, the features of mis-detected frames can be updated to describe the target person, as shown in Fig.~\ref{r-TIAU}. We can also see that the mis-detected frames present lower weights in $T$, indicating their features are suppressed during the aggregation operation and the final video feature is robust to the detection errors.
\begin{figure}[t]
\centering
   \includegraphics[width=0.85\linewidth]{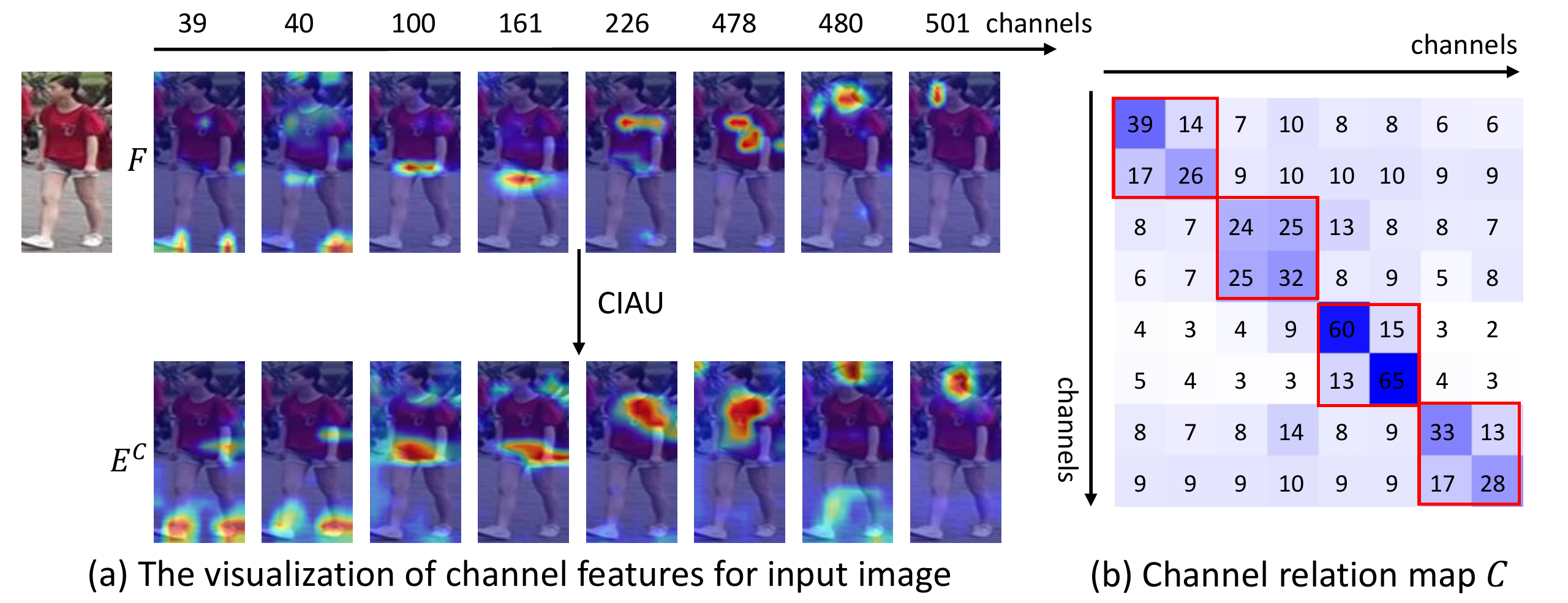}
   \caption{\textbf{Learned Channel Relation Map.} (a) Visualization of the activation maps of initial channel features $F$ and updated channel features $E^C$ by channel IAU for input image. We randomly select 8 channels for clearly visualization. (b) the channel relation maps $C\in\mathbb{R}^{8\times 8}$ among the 8 channels.}
\label{r-CIAU}
\vspace*{-1.5em}
\end{figure}

Fig.~\ref{r-CIAU} visualizes the channel relation map ($C$ in Fig.~\ref{CIAU}). To visualize the channel relation maps, we randomly select 8 channels and visualize their initial features ($F$ in Fig.~\ref{CIAU}), updated features by CIAU ($E^C$ in Fig.~\ref{CIAU}), and the relation maps among the 8 channels. As shown in Fig.~\ref{r-CIAU}, the channel features that focus on the same body parts tend to have a higher correlation. With feature propagation through $C$, each channel can incorporate the specific part information from other channels. As shown in Fig.~\ref{r-CIAU}, the channel feature can focus on more areas of the specific part by CIAU, which enhances its representational power.

\section{Conclusion}
In this paper, we propose an IAU block for globally context modeling that can be effectively implemented by interaction, aggregation, and update operations. The IAU block jointly models spatial-temporal and channel context in a unified framework. We show that by carefully designing the spatial-temporal IAU and channel IAU, the proposed IAUnet achieves state-of-the-art results on both image and video reID tasks over some datasets. In the future, we intend to explore a more advanced metric learning approach to further improve performance. Further, we plan to investigate the use of IAU block beyond person reID and object categorization, such as image and video segmentation.


 

\section*{Acknowledgment}
This work is partially supported by Natural Science Foundation of China (NSFC): 61732004, 61876171 and 61976203, and the University of Chinese Academy of Sciences.

\ifCLASSOPTIONcaptionsoff
  \newpage
\fi

\bibliographystyle{ieeetr}
\bibliography{egbib}

\begin{thebibliography}{10}

\bibitem{multi-channel}
D.~Cheng, Y.~Gong, S.~Zhou, J.~Wang, and N.~Zheng, ``Person re-identification
  by multi-channel parts-based cnn with improved triplet loss function,'' in
  {\em IEEE Conference on Computer Vision and Pattern Recognition}, pp.~1335 --
  1344, 2016.

\bibitem{context-aware}
D.~Li, X.~Chen, Z.~Zhang, and K.~Huang, ``Learning deep context-aware features
  over body and latent parts for person re-identification.,'' in {\em IEEE
  Conference on Computer Vision and Pattern Recognition}, pp.~384--393, 2017.

\bibitem{harmoniou}
W.~Li, X.~Zhu, and S.~Gong, ``Harmonious attention network for person
  re-identification,'' in {\em IEEE Conference on Computer Vision and Pattern
  Recognition}, pp.~2285 -- 2294, 2018.

\bibitem{Hydraplus-net}
X.~Liu, H.~Zhao, M.~Tian, L.~Sheng, J.~Shao, S.~Yi, J.~Yan, and X.~Wang,
  ``Hydraplus-net: Attentive deep features for pedestrian analysis,'' in {\em
  IEEE International Conference on Computer Vision}, pp.~350--359, 2017.

\bibitem{background}
M.~Tian, S.~Yi, H.~Li, S.~Li, X.~Zhang, J.~Shi, J.~Yan, and X.~Wang,
  ``Eliminating background-bias for robust person re-identification,'' in {\em
  IEEE Conference on Computer Vision and Pattern Recognition}, pp.~5794--5803,
  2018.

\bibitem{smoothed}
S.~Bai, X.~Bai, and Q.~Tian, ``Scalable person re-identification on supervised
  smoothed manifold,'' in {\em IEEE Conference on Computer Vision and Pattern
  Recognition}, pp.~2530--2539, 2017.

\bibitem{mars}
L.~Zheng, Z.~Bie, Y.~Sun, J.~Wang, C.~Su, S.~Wang, and Q.~Tian, ``Mars: A video
  benchmark for large-scale person re-identification,'' in {\em European
  Conference on Computer Vision}, pp.~868--884, 2016.

\bibitem{dukereid}
Y.~Wu, Y.~Lin, X.~Dong, Y.~Yan, W.~Quyang, and Y.~Yang, ``Exploit the unknown
  gradually: One-shot video-based person re-identification by stepwise
  learning,'' in {\em IEEE Conference on Computer Vision and Pattern
  Recognition}, pp.~5177--5186, 2018.

\bibitem{zhou2016learning}
B.~Zhou, A.~Khosla, A.~Lapedriza, A.~Oliva, and A.~Torralba, ``Learning deep
  features for discriminative localization,'' in {\em IEEE Conference on
  Computer Vision and Pattern Recognition}, pp.~2921--2929, 2016.

\bibitem{zheng2019re}
M.~Zheng, S.~Karanam, Z.~Wu, and R.~J. Radke, ``Re-identification with
  consistent attentive siamese networks,'' in {\em IEEE Conference on Computer
  Vision and Pattern Recognition}, pp.~5735--5744, 2019.

\bibitem{short-term}
R.~R. Varior, B.~Shuai, J.~Lu, D.~Xu, and G.~Wang, ``A siamese long short-term
  memory architecture for human reidentification.,'' in {\em European
  Conference on Computer Vision}, pp.~135--153, 2016.

\bibitem{3D-convolution}
S.~Ji, W.~Xu, M.~Yang, and K.~Yu, ``3d convolutional neural networks for human
  action recognition,'' {\em IEEE transactions on pattern analysis and machine
  intelligence}, vol.~35, no.~1, pp.~221--231, 2012.

\bibitem{non-local}
X.~Wang, R.~Girshick, A.~Gupta, and K.~He, ``Non-local neural networks,'' in
  {\em IEEE Conference on Computer Vision and Pattern Recognition},
  pp.~7794--7803, 2018.

\bibitem{diversity}
S.~Li, S.~Bak, P.~Carr, C.~Hetang, and X.~Wang., ``Diversity regularized
  spatiotemporal attention for video-based person re-identification,'' in {\em
  IEEE Conference on Computer Vision and Pattern Recognition}, pp.~369--378,
  2018.

\bibitem{RCN}
N.~McLaughlin, J.~M. del Rincon, and P.~C. Miller, ``Recurrent convolutional
  network for video-based person re-identification,'' in {\em IEEE Conference
  on Computer Vision and Pattern Recognition}, pp.~1325--1334, 2016.

\bibitem{IANet}
R.~Hou, B.~Ma, H.~Chang, X.~Gu, S.~Shan, and X.~Chen,
  ``Interaction-and-aggregation network for person re-identification,'' in {\em
  IEEE Conference on Computer Vision and Pattern Recognition}, pp.~9317--9326,
  2019.

\bibitem{residual}
K.~He, X.~Zhang, S.~Ren, and J.~Sun, ``Deep residual learning for image
  recognition,'' in {\em IEEE Conference on Computer Vision and Pattern
  Recognition}, pp.~770 -- 778, 2016.

\bibitem{cifar100}
A.~Krizhevsky and G.~Hinton, ``Learning multiple layers of features from tiny
  images,'' tech. rep., Citeseer, 2009.

\bibitem{Learning}
S.~Paisitkriangkrai, C.~Shen, and A.~van~den Hengel, ``Learning to rank in
  person re-identification with metric ensembles.,'' in {\em IEEE Conference on
  Computer Vision and Pattern Recognition}, pp.~1846--1855, 2015.

\bibitem{stepwise}
Z.~Liu, D.~Wang, and H.~Lu, ``Stepwise metric promotion for unsupervised video
  person re-identification,'' in {\em IEEE International Conference on Computer
  Vision}, pp.~2429--2438, 2017.

\bibitem{hard-aware}
R.~Yu, Z.~Dou, S.~Bai, Z.~Zhang, Y.~Xu, and X.~Bai, ``Hard-aware point-to-set
  deep metric for person re-identification,'' in {\em European Conference on
  Computer Vision}, pp.~188--204, 2018.

\bibitem{gu2019temporal}
X.~Gu, B.~Ma, H.~Chang, S.~Shan, and X.~Chen, ``Temporal knowledge propagation
  for image-to-video person re-identification,'' in {\em IEEE International
  Conference on Computer Vision}, pp.~9647--9656, 2019.

\bibitem{ShiEmbeding}
H.~Shi, Y.~Yang, X.~Zhu, S.~Liao, Z.~Lei, W.~Zheng, and S.~Z. Li, ``Embedding
  deep metric for person re-identification: A study against large variations,''
  in {\em European Conference on Computer Vision}, pp.~732--748, 2016.

\bibitem{Yang2016Large}
Y.~Yang, S.~Liao, L.~Zhen, and S.~Z. Li, ``Large scale similarity learning
  using similar pairs for person verification,'' in {\em AAAI Conference on
  Artificial Intelligence}, 2016.

\bibitem{Rui2013}
Z.~Rui, W.~Ouyang, and X.~Wang, ``Unsupervised graph association for person
  re-identification,'' in {\em IEEE International Conference on Computer
  Vision}, pp.~8321--8330, 2019.

\bibitem{Cuhk}
W.~Li, R.~Zhao, T.~Xiao, and X.~Wang, ``Deepreid: Deep filter pairing neural
  network for person re-identification.,'' in {\em IEEE Conference on Computer
  Vision and Pattern Recognition}, pp.~152--159, 2014.

\bibitem{Triplet}
A.~Hermans, L.~Beyer, and B.~Leibe, ``In defense of the triplet loss for person
  reidentification,'' {\em arXiv preprint arXiv: 1703.07737}, 2017.

\bibitem{zhang2019densely}
Z.~Zhang, C.~Lan, W.~Zeng, and Z.~Chen, ``Densely semantically aligned person
  re-identification,'' in {\em IEEE Conference on Computer Vision and Pattern
  Recognition}, pp.~667--676, 2019.

\bibitem{semantic}
M.~M. Kalayeh, E.~Basaran, M.~Gökmen, M.~E. Kamasak, and M.~Shah, ``Human
  semantic parsing for person re-identification,'' in {\em IEEE Conference on
  Computer Vision and Pattern Recognition}, pp.~1062--1071, 2018.

\bibitem{mask-guided}
C.~Song, Y.~Huang, W.~Ouyang, and L.~Wang, ``Mask-guided contrastive attention
  model for person reidentification,'' in {\em IEEE Conference on Computer
  Vision and Pattern Recognition}, pp.~1179--1188, 2018.

\bibitem{spindle-net}
H.~Zhao, M.~Tian, S.~Sun, J.~Shao, J.~Yan, S.~Yi, X.~Wang, and X.~Tang,
  ``Spindle net: Person re-identification with human body region guided feature
  decomposition and fusion,'' in {\em IEEE Conference on Computer Vision and
  Pattern Recognition}, pp.~1077--1085, 2017.

\bibitem{pose-invariant}
L.~Zheng, Y.~Huang, H.~Lu, and Y.~Yang, ``Pose invariant embedding for deep
  person re-identification.,'' {\em arXiv preprint arXiv:1701.07732}, 2017.

\bibitem{part-aligned}
L.~Zhao, X.~Li, J.~Wang, and Y.~Zhuang, ``Deeply-learned part-aligned
  representations for person re-identification,'' in {\em IEEE International
  Conference on Computer Vision}, pp.~3239 -- 3248, 2017.

\bibitem{PCB}
Y.~Sun, L.~Zheng, Y.~Yang, Q.~Tian, and S.~Wang, ``Beyond part models: Person
  retrieval with refined part pooling (and a strong convolutional baseline),''
  in {\em European Conference on Computer Vision}, pp.~480--496, 2018.

\bibitem{fd}
K.~Zolna, D.~Arpit, D.~Suhubdy, and Y.~Bengio, ``Fraternal dropout,'' {\em
  arXiv preprint arXiv:1711.00066}, 2017.

\bibitem{QAN}
Y.~Liu, J.~Yan, and W.~Ouyang, ``Quality aware network for set to set
  recognition,'' in {\em IEEE Conference on Computer Vision and Pattern
  Recognition}, pp.~4694--4703, 2017.

\bibitem{See}
Z.~Zhou, Y.~Huang, W.~Wang, L.~Wang, and T.~Tan., ``See the forest for the
  trees: Joint spatial and temporal recurrent neural networks for video-based
  person re-identification,'' in {\em IEEE Conference on Computer Vision and
  Pattern Recognition}, pp.~6776--6785, 2017.

\bibitem{jointly}
S.~Xu, Y.~Cheng, K.~Gu, Y.~Yang, S.~Chang, and P.~Zhou, ``Jointly attentive
  spatial-temporal pooling networks for video-based person re-identification,''
  in {\em IEEE International Conference on Computer Vision}, pp.~4743--4752,
  2017.

\bibitem{li2019multi}
J.~Li, S.~Zhang, and T.~Huang, ``Multi-scale 3d convolution network for video
  based person re-identification,'' in {\em AAAI Conference on Artificial
  Intelligence}, vol.~33, pp.~8618--8625, 2019.

\bibitem{HouTCL}
R.~Hou, H.~Chang, B.~Ma, S.~Shan, and X.~Chen, ``Temporal complementary
  learning for video person re-identification,'' in {\em European Conference on
  Computer Vision}, 2020.

\bibitem{Gu3D}
X.~Gu, B.~Ma, H.~Chang, H.~Zhang, and X.~Chen, ``Appearance-preserving 3d
  convolution for video-based person re-identification,'' in {\em European
  Conference on Computer Vision}, 2020.

\bibitem{snippet}
D.~Chen, H.~Li, T.~Xiao, S.~Yi, and X.~Wang, ``Video person re-identification
  with competitive snippet-similarity aggregation and co-attentive snippet
  embedding,'' in {\em IEEE Conference on Computer Vision and Pattern
  Recognition}, pp.~1169--1178, 2018.

\bibitem{Gaze}
X.~He, Y.~Peng, and J.~Zhao, ``Which and how many regions to gaze: Focus
  discriminative regions for fine-grained visual categorization,'' {\em
  International Journal of Computer Vision}, vol.~127, no.~9, pp.~1235--1255,
  2019.

\bibitem{OPA}
Y.~Peng, X.~He, and J.~Zhao, ``Object-part attention model for fine-grained
  image classification,'' {\em IEEE Transactions on Image Processing}, vol.~27,
  no.~3, pp.~1487--1500, 2018.

\bibitem{WSFG}
Y.~Zhang, X.~Wei, J.~Wu, J.~Cai, J.~Lu, V.~A. Nguyen, and M.~N. Do, ``Weakly
  supervised fine-grained categorization with part-based image
  representation.,'' {\em IEEE Transactions on Image Processing}, vol.~25,
  no.~4, pp.~1713--1725, 2016.

\bibitem{WSL}
X.~He and Y.~Peng, ``Weakly supervised learning of part selection model,'' in
  {\em AAAI Conference on Artificial Intelligence}, 2017.

\bibitem{convlstm}
S.~Xingjian, Z.~Chen, H.~Wang, D.~Y. Yeung, W.~K. Wong, and W.~C. Woo,
  ``Convolutional lstm network: A machine learning approach for precipitation
  nowcasting,'' in {\em Advances in neural information processing systems},
  pp.~802--810, 2015.

\bibitem{wang2018videos}
X.~Wang and A.~Gupta, ``Videos as space-time region graphs,'' in {\em European
  Conference on Computer Vision}, pp.~399--417, 2018.

\bibitem{GCT-Gao}
J.~Gao, T.~Zhang, and C.~Xu, ``Graph convolutional tracking,'' in {\em IEEE
  Conference on Computer Vision and Pattern Recognition}, pp.~4649--4659, 2019.

\bibitem{PNGAN}
X.~Qian, Y.~Fu, W.~Wang, T.~Xiang, Y.~Wu, Y.~G. Jiang, and X.~Xue,
  ``Pose-normalized image generation for person re-identification.,'' in {\em
  European Conference on Computer Vision}, pp.~650--667, 2018.

\bibitem{liang2018look}
X.~Liang, K.~Gong, X.~Shen, and L.~Lin, ``Look into person: Joint body parsing
  \& pose estimation network and a new benchmark,'' {\em IEEE Transactions on
  Pattern Analysis and Machine Intelligence}, vol.~41, no.~4, pp.~871--885,
  2018.

\bibitem{Occluded}
S.~Zhang, J.~Yang, and B.~Schiele, ``Occluded pedestrian detection through
  guided attention in cnns,'' in {\em IEEE Conference on Computer Vision and
  Pattern Recognition}, pp.~6995 -- 7003, 2018.

\bibitem{BN}
S.~Ioffe and C.~Szegedy, ``Batch normalization: Accelerating deep network
  training by reducing internal covariate shift,'' {\em arXiv preprint
  arXiv:1502.03167}, 2015.

\bibitem{imagenet}
A.~Karpathy, G.~Toderici, S.~Shetty, T.~Leung, R.~Sukthankar, and L.~Fei-Fei,
  ``Large-scale video classification with convolutional neural networks,'' in
  {\em IEEE Conference on Computer Vision and Pattern Recognition},
  pp.~1725--1732, 2014.

\bibitem{densely}
G.~Huang, Z.~Liu, K.~Q. Weinberger, and L.~van~der Maaten, ``Densely connected
  convolutional networks,'' in {\em IEEE Conference on Computer Vision and
  Pattern Recognition}, pp.~4700--4708, 2016.

\bibitem{SE}
J.~Hu, L.~Shen, and G.~Sun, ``Squeeze-and-excitation networks,'' {\em arXiv
  preprint arXiv:1709.01507}, 2017.

\bibitem{Market1501}
L.~Zheng, L.~Shen, L.~Tian, S.~Wang, J.~Wang, and Q.~Tian, ``Scalable person
  re-identification: A benchmark,'' in {\em IEEE International Conference on
  Computer Vision}, pp.~1116--1124, 2015.

\bibitem{Duke}
Z.~Zheng, L.~Zheng, and Y.~Yang, ``Unlabeled samples generated by gan improve
  the person re-identification baseline in vitro,'' in {\em IEEE International
  Conference on Computer Vision}, pp.~3754--3762, 2017.

\bibitem{msmt17}
L.~Wei, S.~Zhang, W.~Gao, and Q.~Tian, ``Person trasfer gan to bridge domain
  gap for person re-identification,'' in {\em IEEE Conference on Computer
  Vision and Pattern Recognition}, pp.~79--88, 2018.

\bibitem{multicamera}
E.~Ristani, F.~Solera, R.~Zou, R.~Cucchiara, and C.~Tomasi, ``Performance
  measures and a data set for multi-target, multicamera tracking,'' in {\em
  European Conference on Computer Vision}, pp.~17--35, 2016.

\bibitem{LIP}
K.~Gong, X.~Liang, D.~Zhang, X.~Shen, and L.~Lin, ``Look into person:
  Self-supervised structure-sensitive learning and a new benchmark for human
  parsing,'' in {\em IEEE Conference on Computer Vision and Pattern
  Recognition}, pp.~932--940, 2017.

\bibitem{jose2016scalable}
C.~Jose and F.~Fleuret, ``Scalable metric learning via weighted approximate
  rank component analysis,'' in {\em European Conference on Computer Vision},
  pp.~875--890, Springer, 2016.

\bibitem{liao2015person}
S.~Liao, Y.~Hu, X.~Zhu, and S.~Z. Li, ``Person re-identification by local
  maximal occurrence representation and metric learning,'' in {\em IEEE
  Conference on Computer Vision and Pattern Recognition}, pp.~2197--2206, 2015.

\bibitem{SVdnet}
Y.~Sun, L.~Zheng, W.~Deng, and S.~Wang, ``Svdnet for pedestrian retrieval,'' in
  {\em IEEE International Conference on Computer Vision}, pp.~3800--3808, 2017.

\bibitem{cascaded}
Y.~Wang, Z.~Chen, F.~Wu, and G.~Wang, ``Person re-identification with cascaded
  pairwise convolutions,'' in {\em IEEE Conference on Computer Vision and
  Pattern Recognition}, pp.~1470--1478, 2018.

\bibitem{multi-scale-representations}
C.~Yanbei, Z.~Xiatian, and G.~Shaogang, ``Person reidentification by deep
  learning multi-scale representations,'' in {\em IEEE International Conference
  on Computer Vision}, pp.~2590--2600, 2017.

\bibitem{Multi-Level}
X.~Chang, T.~M. Hospedales, and T.~Xiang, ``Multi-level factorisation net for
  person re-identification,'' in {\em IEEE Conference on Computer Vision and
  Pattern Recognition}, pp.~2109--2118, 2018.

\bibitem{KPM}
Y.~Shen, T.~Xiao, H.~Li, S.~Yi, and X.~Wang, ``End-to-end deep
  kronecker-product matching for person re-identification,'' in {\em IEEE
  Conference on Computer Vision and Pattern Recognition}, pp.~6886--6895, 2018.

\bibitem{Mancs}
C.~Wang, Q.~Zhang, C.~Huang, W.~Liu, and X.~Wang, ``Mancs: A multi-task
  attentional network with curriculum sampling for person re-identification,''
  in {\em European Conference on Computer Vision}, pp.~365--381, 2018.

\bibitem{CRF}
D.~Chen, D.~Xu, H.~Li, N.~Sebe, and X.~Wang, ``Group consistent similarity
  learning via deep crf for person re-identification,'' in {\em IEEE Conference
  on Computer Vision and Pattern Recognition}, pp.~8649--8658, 2018.

\bibitem{attention-aware}
J.~Xu, R.~Zhao, F.~Zhu, H.~Wang, and W.~Quyang, ``Attention-aware compositional
  network for person re-identification,'' in {\em IEEE Conference on Computer
  Vision and Pattern Recognition}, pp.~2119--2128, 2018.

\bibitem{pose-sensitive}
M.~S. Sarfraz, A.~Schumann, A.~Eberle, and R.~Stiefelhagen, ``A pose-sensitive
  embedding for person re-identification with expanded cross neighborhood
  re-ranking,'' in {\em IEEE Conference on Computer Vision and Pattern
  Recognition}, pp.~420--429, 2018.

\bibitem{zhong2017random}
Z.~Zhong, L.~Zheng, G.~Kang, S.~Li, and Y.~Yang, ``Random erasing data
  augmentation,'' {\em arXiv preprint arXiv:1708.04896}, 2017.

\bibitem{adam}
D.~P. Kingma and J.~Ba, ``Adam: A method for stochastic optimization,'' {\em
  arXiv preprint arXiv:1412.6980}, 2014.

\bibitem{multi-scale}
Y.~Chen, X.~Zhu, and S.~Gong, ``Person re-identification by deep learning
  multi-scale representations,'' in {\em IEEE International Conference on
  Computer Vision}, pp.~2590--2600, 2017.

\bibitem{Cross-view}
H.~X. Yu, A.~Wu, and W.~S. Zhen, ``Cross-view asymmetric metric learning for
  unsupervised person re-identification,'' in {\em IEEE International
  Conference on Computer Vision}, pp.~994--1002, 2017.

\bibitem{going}
C.~Szegedy, W.~Liu, Y.~Jia, P.~Sermanet, S.~Reed, D.~Anguelov, D.~Erhan,
  V.~Vanhoucke, and A.~Rabinovich, ``Going deeper with convolutions,'' in {\em
  IEEE Conference on Computer Vision and Pattern Recognition}, pp.~1--9, 2015.

\bibitem{pose-driven}
C.~Su, J.~Li, S.~Zhang, J.~Xing, W.~Gao, and Q.~Tian, ``Pose-driven deep
  convolutional model for person re-identification,'' {\em arXiv preprint
  arXiv:1709.08325}, 2017.

\bibitem{Glad}
L.~Wei, S.~Zhang, H.~Yao, W.~Gao, and Q.~Tian, ``Glad: global-local-alignment
  descriptor for pedestrian retrieval,'' in {\em Proceedings of ACM
  international conference on Multimedia}, pp.~420--428, 2017.

\bibitem{re-rank}
Z.~Zhong, L.~Zheng, D.~Cao, and S.~Li, ``Re-ranking person re-identification
  with k-reciprocal encoding.,'' in {\em IEEE Conference on Computer Vision and
  Pattern Recognition}, pp.~1318--1327, 2017.

\bibitem{sequence-decision}
J.~Zhang, N.~Wang, and L.~Zhang, ``Multi-shot pedestrian re-identification via
  sequential decision making,'' in {\em IEEE Conference on Computer Vision and
  Pattern Recognition}, pp.~6781--6789, 2018.

\bibitem{RQAN}
G.~Song, B.~Leng, Y.~Liu, C.~Hetang, and S.~Cai, ``Region-based quality
  estimation network for large-scale person re-identification,'' {\em arXiv
  preprint arXiv:1711.08766}, 2017.

\bibitem{zhao2019attribute}
Y.~Zhao, X.~Shen, Z.~Jin, H.~Lu, and X.-s. Hua, ``Attribute-driven feature
  disentangling and temporal aggregation for video person re-identification,''
  in {\em IEEE Conference on Computer Vision and Pattern Recognition},
  pp.~4913--4922, 2019.

\bibitem{VRSTC}
R.~Hou, B.~Ma, H.~Chang, X.~Gu, S.~Shan, and X.~Chen, ``Vrstc: Occlusion-free
  video person re-identification,'' in {\em IEEE Conference on Computer Vision
  and Pattern Recognition}, pp.~7183--7192, 2019.

\bibitem{identity}
K.~He, X.~Zhang, S.~Ren, and J.~Sun, ``Identity mappings in deep residual
  networks,'' in {\em European Conference on Computer Vision}, pp.~630--645,
  2016.

\bibitem{Aggregated}
S.~Xie, R.~Girshick, P.~Dollar, Z.~Tu, and K.~He, ``Aggregated residual
  transformations for deep neural networks,'' in {\em IEEE Conference on
  Computer Vision and Pattern Recognition}, pp.~1492--1500, 2016.

\bibitem{Dualnet}
S.~Hou, X.~Liu, and Z.~Wang, ``Dualnet: Learn complementary features for image
  recognition,'' in {\em IEEE Conference on Computer Vision and Pattern
  Recognition}, pp.~502--510, 2017.

\bibitem{CAMA}
W.~Yang, H.~Huang, Z.~Zhang, X.~Chen, K.~Huang, and S.~Zhang., ``Towards rich
  feature discovery with class activation maps augmentation for person
  re-identification.,'' in {\em IEEE Conference on Computer Vision and Pattern
  Recognition}, pp.~1389--1398, 2019.

\end{thebibliography}

%







\end{document}